\documentclass[final]{cvpr}
\usepackage{times}
\usepackage{epsfig}
\usepackage{graphicx}
\usepackage{amsmath}
\usepackage{amssymb}
\usepackage{dsfont}
\usepackage{booktabs}
\usepackage{tabularx}
\usepackage{arydshln}
\usepackage[pagebackref=true,breaklinks=true,colorlinks,bookmarks=false]{hyperref}

\usepackage[ruled,vlined]{algorithm2e}

\def\cvprPaperID{2069} % *** Enter the CVPR Paper ID here
\def\confYear{CVPR 2021}

\begin{document}

%%%%%%%%% TITLE
\title{Stay Positive: Non-Negative Image Synthesis for Augmented Reality}

% \author{
% Katie Luo$^{1}$\thanks{Equal contribution.}\ , 
% Guandao Yang$^{1,2}$\footnotemark[1], 
% Wenqi Xian$^{1,2}$, 
% Harald Haraldsson$^3$, 
% Bharath Hariharan$^1$, 
% Serge Belongie$^{1,2}$\\
% $^1$Cornell University 
% \hspace{15pt}
% $^2$Cornell Tech
% }

\author{
Katie Luo\thanks{Denotes equal contribution.} $^{,1}$\hspace{10pt}
Guandao Yang\footnotemark[1] $^{,1,2}$\hspace{10pt}
Wenqi Xian$^{1,2}$\hspace{10pt}
Harald Haraldsson$^{2}$\\
% \hspace{10pt}
Bharath Hariharan$^{1}$\hspace{10pt}
Serge Belongie$^{3}$\\ 
$^1$Cornell University\hspace{14pt}$^2$Cornell Tech\hspace{14pt}$^3$University of Copenhagen
}

\definecolor{mypink}{cmyk}{0, 0.7808, 0.4429, 0.1412}

\newcommand{\todo}[1]{\textcolor{red}{todo:#1}}
\newcommand{\wenqi}[1]{\textcolor{mypink}{wenqi:#1}}
\newcommand{\guandao}[1]{\textcolor{blue}{guandao:#1}}
\newcommand{\katie}[1]{\textcolor{green}{katie:#1}}
\newcommand{\bh}[1]{\textcolor{cyan}{BH:#1}}
\newcommand{\sjb}[1]{\textcolor{magenta}{SJB:#1}}
\newcommand{\argmin}{\mathop{\mathrm{arg\,min}}}
\newcommand{\argmax}{\mathop{\mathrm{arg\,max}}}
\maketitle

%%%%%%%%% ABSTRACT
\begin{abstract}
In applications such as optical see-through and projector augmented reality, producing images amounts to solving non-negative image generation, where one can only add light to an existing image.
Most image generation methods, however, are ill-suited to this problem setting, as they make the assumption that one can assign arbitrary color to each pixel.
In fact, naive application of existing methods fails 
%catastrophically
even in simple domains such as MNIST digits, since one cannot create darker pixels by adding light.
We know, however, that the human visual system can be fooled by optical illusions involving certain spatial configurations of brightness and contrast.
Our key insight is that one can leverage this behavior to produce high quality images with negligible artifacts.
For example, we can create the illusion of darker patches by brightening surrounding pixels.
We propose a novel optimization procedure to produce images that satisfy both semantic and non-negativity constraints.
Our approach can incorporate existing state-of-the-art methods, and exhibits strong performance in a variety of tasks including image-to-image translation and style transfer.

\end{abstract}

%%%%%%%%% BODY TEXT
\section{Introduction}\label{sec:intro}

\begin{figure}[t]
    \centering
    \includegraphics[width=\linewidth]{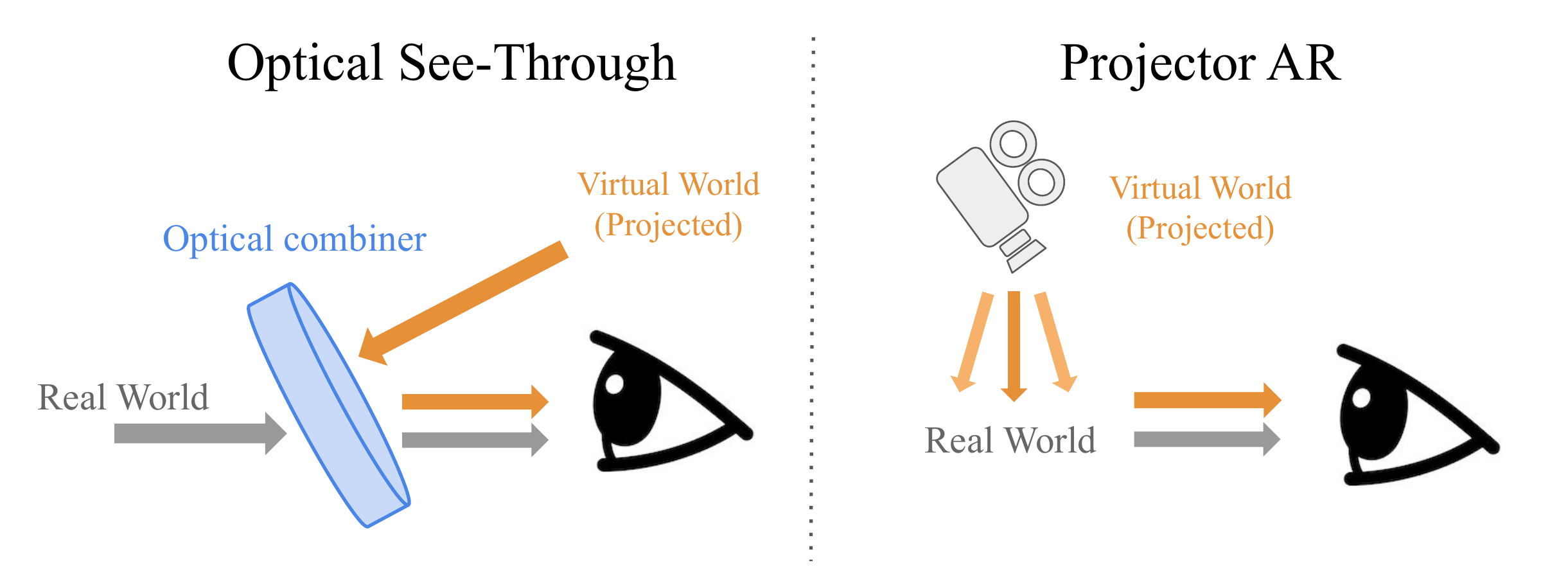}
    \caption{Illustration of the image formation process in augmented reality settings.
    Left: optical see-through setting. Right: projector AR setting. 
    In both settings, the image is created by adding light to existing light sources from the real world, which motivates the non-negative image generation problem.
    }
    \label{fig:prob-set-up-AR}
\end{figure}

The design of images that combine views of the real world with aligned, overlaid content remains a key challenge on the path toward widespread adoption of augmented reality devices.
Consider the problem of image generation in an optical see-through (OST) or projector AR device, where the image is formed by combining light from the real world with light generated by a projector; see Fig.~\ref{fig:prob-set-up-AR}. 
Despite the success of modern image generation methods in applications including deep image generation ~\cite{karras2019style, karras2019analyzing, karras2017progressive}, image synthesis~\cite{Park_2019_CVPR, Wang_2018_CVPR},  
image-to-image translation~\cite{zhu2017unpaired,liu2017unsupervised,Huang_2018_ECCV}, and style transfer~\cite{huang2017arbitrary, sheng2018avatar},
all of the aforementioned methods assume full control over each pixel's color.
This assumption, however, is not realistic in the OST or projector AR setting where one can add, but not subtract, light. 
In fact, naively applying any of the current state-of-the-art image generation methods will result in low-quality images in conflict with the underlying image, often riddled with ghosting artifacts. 

In this work, we focus on the task of image synthesis under a non-negativity constraint, which limits the types of images that can be generated.
Since we restrict ourselves to adding a non-negative residual to the input image, we would in theory only be able to generate images in our half-space. 
Intuitively, one cannot add black-framed glasses onto Brad Pitt's face, since doing so would require us to attenuate the light in the desired area, \ie, adding negative light, which is not physically possible. 
With such limitations, current hardware solutions compromise by relying on dimming the incoming light instead, using methods such as a sunglasses-like effect for OST and a darkened room for projector AR. 
In fact, current solutions use very dark tints, admitting only $15-20$\% \cite{otsheadmountlimits,brightview} of the light from the original scene. 
While this delivers increased control over the pixel dynamic range, 
blocking too much light in the scene limits the kind of settings in which OST or projector AR can be applied.

{
\renewcommand{\tabcolsep}{1pt}
\begin{figure}
\begin{center}
\begin{tabular}{cc}
\includegraphics[width=0.48\linewidth]{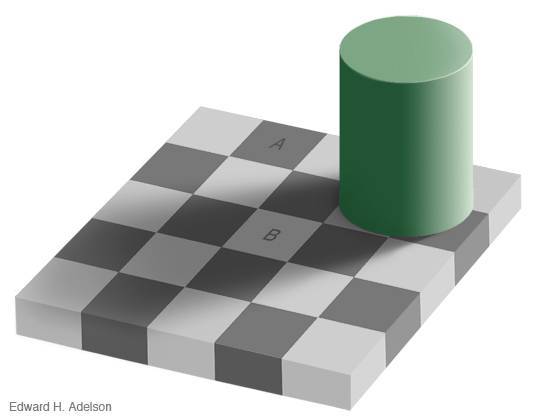} &
\includegraphics[width=0.48\linewidth]{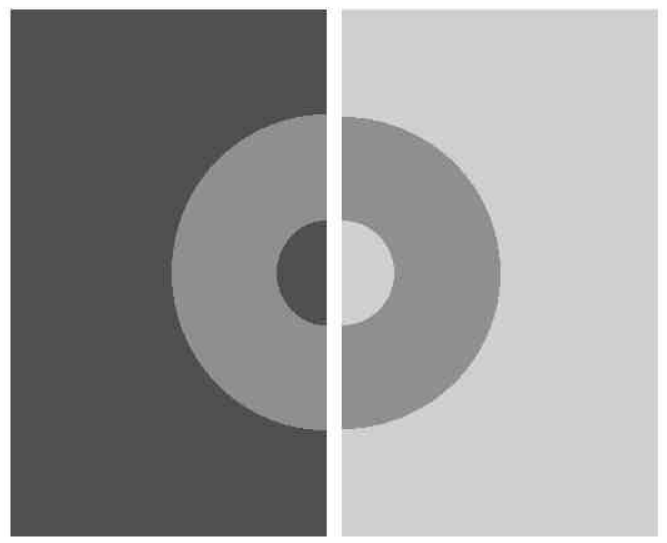}
\end{tabular}
\end{center}
\caption{Examples of lightness illusions. Left: the Checker-shadow illusion ~\cite{adelson200024}. Areas $A$ and $B$ have the same color, yet $B$ appears brighter.
Right: Koffka ring illusion~\cite{koffka2013principles}. 
Both half-rings have the same color, but the right one looks darker.
These two examples illustrated that human do not perceive brightness in an absolute way.
}
\label{fig:illusions}
\end{figure}
}

Without the ability to subtract light or dim the scene, is it still possible to succeed in image-to-image translation tasks such as turning a day image into a night image?
If humans perceived luminance and chrominance in an objective manner, then this would be virtually impossible.
Fortunately, the human visual system is tolerant to a wide range of photometric deviations from a ``true'' image \cite{visualinfostimulus,visualsizecontrastillusion},
as illustrated by popular optical illusions that trick us into seeing colors or relative brightness orderings that aren't there (\eg Figure~\ref{fig:illusions}). 
These quirks offer us a path to a solution for the challenging task of non-negative image generation: we can leverage properties of the human visual system to enlarge the solution space.
For example, \textit{lightness constancy}~\cite{adelson200024} suggests that humans do not perceive light in an absolute way, but instead take into account many factors such as geometry, texture, and surrounding neighborhoods.
We may therefore no longer need to block light to create the perception of a darker pixel -- we could instead add light to its peripheral locations.
Another quirk of human vision is that we are relatively insensitive to visual glitches when we are not paying attention to them~\cite{Osberger1999PerceptualVM}.
This allows us to introduce errors in some non-salient locations without being noticed.

In this paper, we approach the novel task of non-negative image generation by designing a framework that can take advantage of both existing image-to-image translation methods and human visual perception quirks. 
Specifically, we first incorporate existing image-to-image translation models to generate a proposal image that satisfies the task, but is not necessarily feasible in the non-negative generation setting.
We then finetune the proposal image so that it can be produced by adding non-negative light to the input image while remaining perceptually unchanged.
Our method is compatible with almost all existing image-to-image translation methods,
%which can be used in our first step,
thereby permitting its application to a wide variety of tasks.
% versatile
While we mainly appeal to \textit{lightness constancy} in this paper, our framework can be extended to model other types of perceptual illusions, which we leave to future explorations.
We empirically show that our method outperforms all strong baselines that naively adapt state-of-the-art models on this task, both quantitatively and qualitatively, even in settings when much more light is allowed in the optical combiner.
% combinator.
Finally, we provide detailed analysis of how -- and when -- our method works. 
%to provide insight about how our method is capable of leveraging human perceptual illusions to produce photo-realistic images. 

\section{Non-negative Image Generation}\label{sec:non-negative-img-gen}

We now specify the non-negative image generation problem, in which the model can only add light to the input image when producing the output image. 
We narrow our scope to the specific problem of producing an image that fulfills a certain semantic or style category, as opposed to any arbitrary image. 
This is motivated by the idea that AR users are more likely to specify a category instead of a detailed image. 

During training, one is provided with samples from $P_X$ and $P_Y$, the distributions of the images of the input domain and target domain, respectively.
The goal is to learn a model that can take an input image $x\sim P_X$ and produce a non-negative residual $R_\theta(x)$ that can be combined with the input to produce an image in $P_Y$. 
Formally, non-negative image generation amounts to solving the following constrained optimization problem:
\begin{equation}
    \begin{aligned}
        \argmax_\theta&\quad \mathbb{E}_{x \sim P_X}\left[\log P_Y(\alpha x + \beta R_\theta(x))\right] \\
        \text{s.t.}&\quad \forall i,j,~ 0 \leq R_\theta(x)_{i,j} \leq 1,
    \end{aligned}\label{eq:general-pb}
\end{equation}
where $\alpha, \beta \in [0, 1]$ are parameters controlling influence from the input image and the predicted residual, respectively, and (with a slight abuse of notation) $R_\theta(x)_{i,j}$ denotes the $i$- by $j$-th pixel of residual $R_\theta(x)$.

While the problem formulation in Equation~\ref{eq:general-pb} can be used in a number of AR settings, we focus on the optical see-through (OST) case.
Here, $\alpha$ represents the amount of light transmitted through the semi-transparent mirror.
At the same time, the amount of light reflected from the semi-transparent mirror is $1 - \alpha$, so the upper bound is $\beta=1-\alpha$.
We refer to this simplified optical formulation for the rest of the work.
Note that under this formulation, when $\alpha=0$, the problem reduces to the video see-through (VST) setting, \ie, the traditional unconstrained version.
The higher the value of $\alpha$, the harder the problem becomes, since the amount of light at the disposal of the generator $R_\theta$ becomes more limited.

\begin{figure*}[t]
    \centering
    \includegraphics[width=\linewidth,trim={0 0 0 0.7},clip]{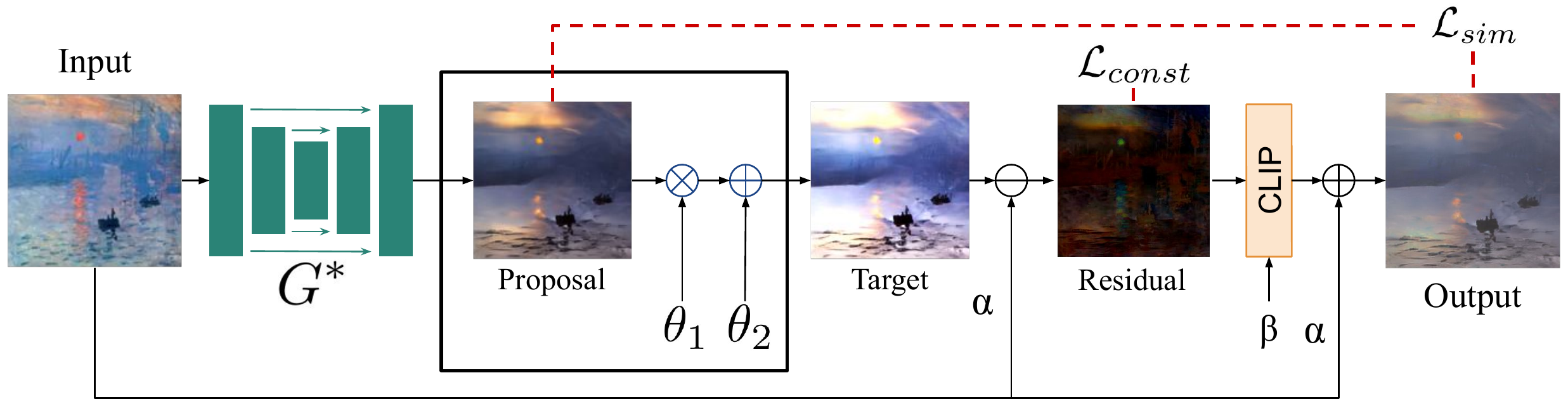}
    \caption{\textbf{Illustration of our two-stage training pipeline.} 
    Our framework's overall schematic consists of the image proposal step and semantic-preserving generator step for the residual. The final output is an optical combination between the input and residual, clipped to be physically feasible. We rely on very few parameters for efficient training.}
    \label{fig:model}
\end{figure*}

\section{Method}\label{sec:method}
In developing our proposed method, we aim to draw upon the strengths of state-of-the-art image generation methods while availing ourselves of degrees of freedom provided by the human visual system.

%%%%%%%%%%%%%%%%%%%%%%%%%%%%%%%%%%%%%%%%%%%%%%%%%%%%%%%%%%%%%%%%%%%%%%%%
% Separation
%%%%%%%%%%%%%%%%%%%%%%%%%%%%%%%%%%%%%%%%%%%%%%%%%%%%%%%%%%%%%%%%%%%%%%%%
\subsection{A two stage approach}\label{sec:two-step}

To make a model compatible with existing image generators and adaptable to quirks of human visual perception, we propose to break down the non-negative image generation problem into two steps.
The first step is to solve the image generation problem without the non-negativity constraint.
%This allows us to leverage the advantages from existing and future methods in the field of image-to-image translation.
In the second step, we finetune the image produced by the first step with the aim of making it perceptually unchanged under the non-negativity constraint.
%In this step, we will design objectives and models to leverage the visual perceptual properties to enlarge the solution space where we might find a desirable solution that's feasible to the constraint.

Formally, we assume that we can produce a model $G^*$ that takes an image from domain $X$ and outputs one that is in domain $Y$ while preserving the structure of the input images. Specifically, we take a model $G^*$ that minimizes the following objective:
\begin{align}
    \min_G\mathbb{E}_{x\sim \mathcal{X} } \left[\mathcal{S}(G(x), x)-\log{P_Y(G(x))}\right] \label{eq:semantic-stage},
\end{align}
where $\mathcal{S}(\cdot, \cdot)$ returns the structural similarity between two images.
Most image-to-image translation works such as Pix2Pix~\cite{isola2017image} and CycleGAN~\cite{zhu2017unpaired} can achieve this step. 

At this point, we can produce an image proposal $y=G^*(x)$ for any input image $x$.
The image proposal $y$ is a good solution as long as it can be feasibly produced under the non-negativity constraints imposed by Equation~\ref{eq:general-pb}.
The end goal, however, does not require the output image to be exactly the same as $y$ --- in fact, any image that is perceptually indistinguishable from $y$ should suffice.
Therefore, the second step can be formulated as finding a target image such that it is perceptually similar to input image $x$ while being physically realizable in the non-negative image generation regime.
Formally, for each pair of input image $x$ and image proposal $y$, we want to find an image $F_{\boldsymbol{\theta}}(x, y)$ that optimizes the following objective and constraints:
\begin{equation}
    \begin{aligned}
        \min_{\boldsymbol{\theta}}&\quad
        \mathcal{L}_{sim}(F_{\boldsymbol{\theta}}(x, y), y) \\
        \text{s.t. }&\quad \forall i,j, ~0 \leq F_{\boldsymbol{\theta}}(x, y)_{i,j} - \alpha x_{i,j} \leq \beta,  
    \end{aligned}\label{eq:optic-stage}
\end{equation}
% Techniques
where $\mathcal{L}_{sim}(\cdot, \cdot)$ measures the perceptual similarity between two images.
%Formulating the problem in such way allows us to leverage the fact that human visual systems are tolerance to different mistakes in the images. 
For example, if $y'$ represents $y$ altered by perturbations undetectable to the human eye, then $\mathcal{L}_{sim}(y', y)=0$. 
Prior works have studied measuring perceptual similarity for a variety of applications including lossy image compression~\cite{mier2021deep, reisenhofer2018haar} and image generation~\cite{Zhang2018TheUE}.
In the following sections, we will show how to design $\mathcal{L}_{sim}$ to encode lightness constancy, and how to solve this problem by optimizing a few parameters using stochastic gradient descent.
% In the following section, we will show that how to solve this problem by designing the visual perceptual similarity loss, fulfilling the constraints, as well as predicting a residual with minimum number of parameters.
% Note that we are solving an approximate minimization to the problem objective in Equation~\ref{eq:general-pb}, as setting $G^*$ beforehand reduces our search-space.
% , and $G_\theta(x)$ might be the one that is infeasible to produce. 

%%%%%%%%%%%%%%%%%%%%%%%%%%%%%%%%%%%%%%%%%%%%%%%%%%%%%%%%%%%%%%%%%%%%%%%%
% Perceptual similarity
%%%%%%%%%%%%%%%%%%%%%%%%%%%%%%%%%%%%%%%%%%%%%%%%%%%%%%%%%%%%%%%%%%%%%%%%
\subsection{Perceptual similarity}\label{sec:reconloss}

{\renewcommand{\tabcolsep}{2pt}
\begin{figure*}
   \centering
\begin{tabular}{c:c@{}c:c@{}c}
$O_{\boldsymbol{\theta}}(x,y)$ & No normalization & Minima & With normalization & L2-Normed Minima  \\
% \hline\hline

\includegraphics[width=0.19\textwidth]{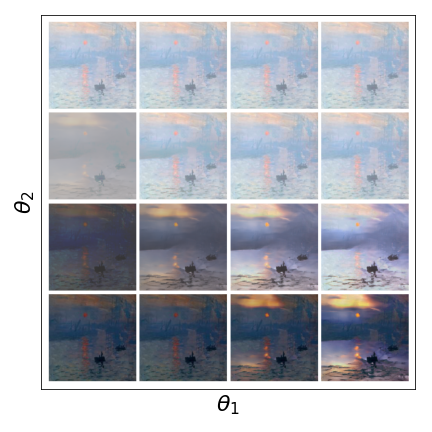}&
\includegraphics[width=0.19\textwidth]{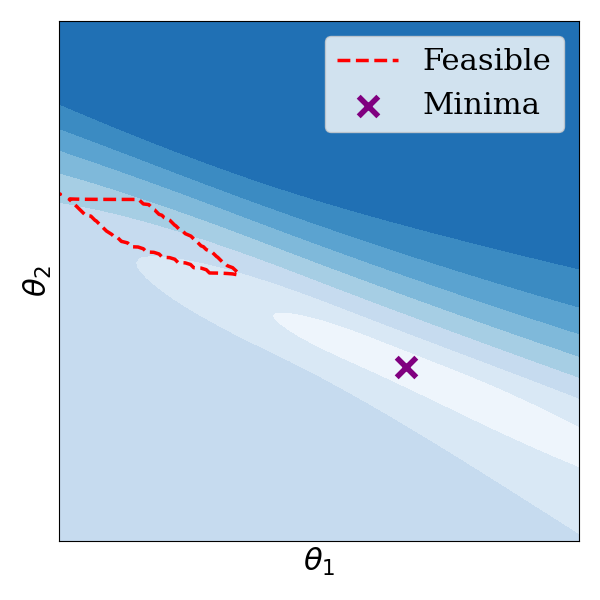}&
\includegraphics[width=0.19\textwidth]{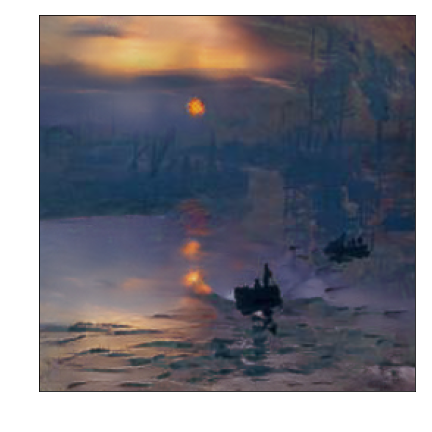}&
\includegraphics[width=0.19\textwidth]{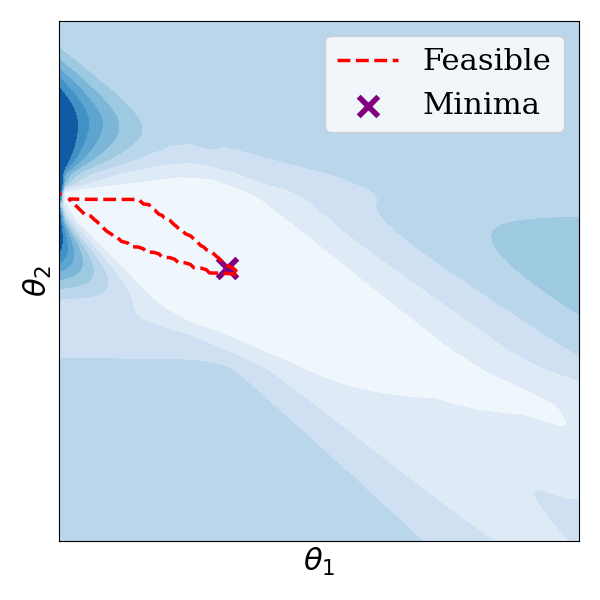}&
\includegraphics[width=0.19\textwidth]{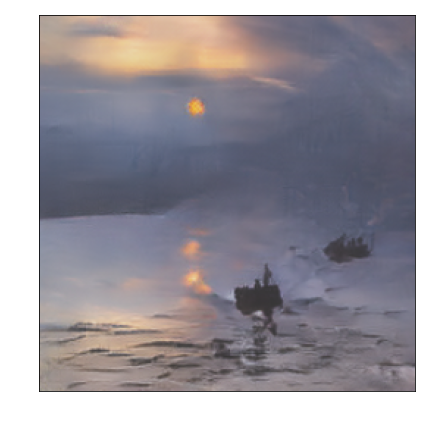}
\end{tabular}
\vspace{0.5em}
\caption{
Visualizations of our Semantic-Preserving Generator outputs (Left). 
By varying the $\boldsymbol{\theta}$ parameters, we can control the brightness and contrast of the final output image. 
We visualize the Perceptual Similarity loss with and without normalization (Middle and Right respectively), and outline the region of parameters corresponding to physically realizable residuals in \textcolor{red}{red}. 
Darker color indicates larger loss.
The output image of the minimum of the two losses are shown to the right of the loss contour. 
Observe that normalizing the images accomplishes two things: 1. the loss contour changes to include the physically feasible region, and 2. the perceptual quality of the image is improved and the output exhibits less ``ghosting" artifacts from the input image. 
% \todo{two points: 1. normalization makes the thing feasible; 2. normalizxation creates less artifact.}
% Observe the minima corresponding to the loss with normalization encloses the physical feasibility region, corresponds better with visual quality
}
\label{fig:perceptual-similarity} % I can do without the label too
\end{figure*}
}

%The objective $\mathcal{L}_{sim}$ determines how close our model considers two images to be.
The simplest (and strictest) formulation for $\mathcal{L}_{sim}$ is that two images are the same only if the pixel values match everywhere (\ie, $\mathcal{L}_{sim}(a, b)= \|a - b\|^2$).
%As mentioned in Section~\ref{sec:intro}, such formulation is problematic as it does not take into consideration human perception properties such as \textit{lightness constancy}. \katie{not sure if we want to italicise all instances of this...}
We instead propose to relax this by incorporating the notion of lightness constancy.
Formally, we define the similarity metric as:
\begin{align}
    \mathcal{L}_{sim}(a, b) = \|N(a) - N(b)\|^2\label{eq:perc-sim-norm},
\end{align}
where $N(\cdot)$ is a normalization function that brings two perceptually similar images closer together.
For example, to capture the idea that humans are visually insensitive to global lightness changes, which in part explains lightness constancy, we normalize the dynamic range the image by using Color Histogram Stretching~\cite{Wang2011FastAW}:
\begin{align}
    N(x) = \frac{x - x_{min}}{x_{max} - x_{min}}\label{eq:range-norm},
\end{align}
where $x_{min}$ and $x_{max}$ are the smallest and largest pixel values in image $x$, respectively.
For RGB images, we apply such normalization independently to all three channels.
With this operation in place, adding a constant to every pixel in the image will not affect the output since $N(ax+b)=N(x)$.
While Equation~\ref{eq:perc-sim-norm} is not restricted to just one normalization method or just one notion of perceptual quality, we will focus on the lightness constancy property and will assume the use of Equation~\ref{eq:range-norm} unless specified otherwise.
Figure~\ref{fig:perceptual-similarity} shows how loss with normalization leads to feasible solutions and outputs with less artifacts.

%%%%%%%%%%%%%%%%%%%%%%%%%%%%%%%%%%%%%%%%%%%%%%%%%%%%%%%%%%%%%%%%%%%%%%%%
% Constrain loss
%%%%%%%%%%%%%%%%%%%%%%%%%%%%%%%%%%%%%%%%%%%%%%%%%%%%%%%%%%%%%%%%%%%%%%%%
\subsection{Soft Constraint Loss}\label{sec:resloss}

The above expression of perceptual similarity with normalization makes the constrained optimization problem in Equation~\ref{eq:optic-stage} non-convex, and thereby a more more challenging problem.
% The leniency of human visual perception system, again, comes into rescue.
As mentioned in Section~\ref{sec:intro}, we are not sensitive to minor changes in color, especially in high texture regions of a natural image.
This allows our system to make some mistakes by making the target $F_{\boldsymbol{\theta}}(x, y)$ not \textit{completely} physically realizable at all pixels, as long as the error between the actual output and the target output is relatively small.
% This allow our system to make some mistakes by aiming to generate images that are not \textit{completely} physically realizable in all pixels, as long as the error made in those broken pixels is relatively small.
Thus, we relax the hard constraint on the residual $r$ into minimizing the following soft constraint:
\begin{equation}
\begin{aligned}
    \mathcal{L}_{const}(r, a, b) = \gamma \sum_{i,j}|\max(\min(r_{i,j}, b), a) - r_{i,j}|,
\end{aligned}
\end{equation}
where $\gamma > 0$ is a hyper-parameter that controls the trade-off between perceptual similarity and residual constraint satisfaction.
This soft constraint loss gives a continuous penalty to places where the pixel is not physically realizable, and it is differentiable everywhere except at $a$ or $b$.
By using the soft constraint as a loss function, we get the added benefit of penalizing the values depending on how infeasible they are. 
This allows the model to balance between producing a strictly physically realizable image and producing an image that is perceptually similar to the image proposal, albeit containing a few errors.
% \todo{figure to show the claim?}

%%%%%%%%%%%%%%%%%%%%%%%%%%%%%%%%%%%%%%%%%%%%%%%%%%%%%%%%%%%%%%%%%%%%%%%%
% Structure preserving generator
%%%%%%%%%%%%%%%%%%%%%%%%%%%%%%%%%%%%%%%%%%%%%%%%%%%%%%%%%%%%%%%%%%%%%%%%

\begin{algorithm}[t]
\SetAlgoLined
\KwResult{Output image $O_{\boldsymbol{\theta}}(x,y)$}
 \textbf{Input}: Device parameters $\alpha$, $\beta$, image $x$, proposal $y$, learning rate $lr$
 \;
 Initialize $\boldsymbol{\theta} = [1, 0]$\;
 \While{not converged}{
  $F_{\boldsymbol{\theta}}(x,y) = \theta_1 y + \theta_2$\;
  $r = F_{\boldsymbol{\theta}}(x,y) - \alpha x$\;
  $O_{\boldsymbol{\theta}}(x, y) = min(max(r, 0), \beta) + \alpha x$\;
  $\mathcal{L}({\boldsymbol{\theta}}) \leftarrow \mathcal{L}_{sim}(O_{\boldsymbol{\theta}}(x, y), y) + \mathcal{L}_{const}(r, 0, \beta)$\;
  ${\boldsymbol{\theta}} \leftarrow Adam(\nabla_{\boldsymbol{\theta}} \mathcal{L}({\boldsymbol{\theta}}), lr)$\;
 }
 \caption{Semantics-Preserving Generator}
 \label{alg:const-gen}
\end{algorithm}

\subsection{Semantics-Preserving Generator}\label{sec:offset}
\newcolumntype{Y}{>{\centering\arraybackslash}X}
\newcolumntype{L}{>{\arraybackslash}X}
\begin{table*}[t]
\begin{center}
\begin{tabularx}{\textwidth}{l*{10}{Y}}
\toprule
       & \multicolumn{3}{c}{Satellite $\rightarrow$ Map \cite{isola2017image}}                                & \multicolumn{3}{c}{Map $\rightarrow$ Satellite \cite{isola2017image}}                                & \multicolumn{3}{c}{Day $\rightarrow$ Night \cite{Laffont14}}                                    \\
\cmidrule(lr){2-4} \cmidrule(lr){5-7} \cmidrule(lr){8-10} 
Method & 
\multicolumn{1}{c}{FID($\downarrow$)} & \multicolumn{1}{c}{KID($\downarrow$)} & \multicolumn{1}{c}{N-PSNR($\uparrow$)} & \multicolumn{1}{c}{FID($\downarrow$)} & \multicolumn{1}{c}{KID($\downarrow$)} & \multicolumn{1}{c}{N-PSNR($\uparrow$)} & \multicolumn{1}{c}{FID($\downarrow$)} & \multicolumn{1}{c}{KID($\downarrow$)} & \multicolumn{1}{c}{N-PSNR($\uparrow$)} \\ 
\midrule
Heuristic     & 252.33                  & 19.03                  & 20.16                      & 200.96                  & 18.08                  & 33.07                      & 134.97                  & 4.53  & 39.00                      \\
From scratch     & 268.79                  & 19.38                  & 26.19                      & 305.82                  & 33.29                  & 30.63                      & 218.09                  &  11.75     & 27.67                      \\
Finetuning     & 254.86                  & \textbf{16.22}                  & \textbf{30.77}                      & 265.87                  & 24.97                  & 34.63                      & 195.14                  & 11.86                    & 29.00                      \\
\midrule
Ours   & \textbf{238.93}         & 17.90         & 23.78             & \textbf{125.75}         & \textbf{12.76}         & \textbf{68.90}             & \textbf{110.83}                  & \textbf{3.13}                    & \textbf{47.42}                      \\ 
\bottomrule
\end{tabularx}
\end{center}
\vspace{-0.5em}
\caption{Method performances on Map$\leftrightarrow$Satellite and Day$\rightarrow$Night datasets. }
\label{tab:baseline}
\end{table*} 

The final piece of the puzzle is the design of the generator $F_{\boldsymbol{\theta}}(x, y)$.
$F_{\boldsymbol{\theta}}$ takes the input image $x$ and the semantic prototype $y$ and aims to predict a target image that looks similar to $y$ and is also physically realizable by adding light to $\alpha x$.
The naive solution for $F_{\boldsymbol{\theta}}(x, y)$ is to return $y$ directly, but such solution is likely to break the constraints since the brightness of the $X$ domain could be drastically different from that of the $Y$ domain.
In the other extreme, we can use a universal approximator like a neural network to approximate the function (\ie, predicting colors of each pixel).
Since such a generator imposes almost no domain specific knowledge, it contains lots of parameters and can be empirically difficult to train. 
In this paper, we propose a semantics-preserving generator that is flexible enough to include a physically feasible solution while building upon $y$ as a very good starting point.

Specifically, we want the generator to be able to explore images that preserve the structure of the semantic image $y$ in order to find an image that best respects the physical constraints.
Note an affine transformation on an image $y$ into $\theta_1 y+\theta_2$ does not change the perceptual loss $\mathcal{L}_{sim}$ since $N(\theta_1 y + \theta_2) = N(y)$ for any scalar $\theta_1$ and $\theta_2$.
This reflects the assumption that humans have the ability to do white balancing and adjust to different contrasts.
One can potentially find a physically realizable image by choosing $\boldsymbol{\theta}=(\theta_1, \theta_2)$.
% Formally, let $g_\theta(x, y)$ and $h_\theta(x,y)$ be a function that outputs parameters to preserve semantic of $y$. 
Formally, we define the target image $F_{\boldsymbol{\theta}}(x, y)$ and the physically realizable output image $O_{\boldsymbol{\theta}}(x, y)$ as follow: 
\begin{equation}
\begin{aligned}
F_{\boldsymbol{\theta}}(x, y) &= \theta_1 y + \theta_2, \\
O_{\boldsymbol{\theta}}(x, y) &= \max(\min(F_{\boldsymbol{\theta}}(x, y) - \alpha x, \beta), 0) + \alpha x.
\end{aligned}    
\end{equation}
% where for each $i$- by $j$-th pixel. Note that with this generator, arbitrary $g_\theta(x, y)$ and $h_\theta(x,y)$ will not result in a physically realizable residual. 
% Under these cases, we must clip the values to be within the feasible range. We explore different structures for $g_\theta(x, y)$ and $h_\theta(x,y)$ in the Section~\ref{sec:experiments}.
We adjust the final objective to also minimize the perceptual similarity between the actual output image $O_{\boldsymbol{\theta}}$ and the image proposal $y$.
This allows $\mathcal{L}_{sim}$ to also provide training signal, which empirically leads to better performance.
Thus, the final training objective is:
\begin{equation}
    \begin{aligned}
    \mathcal{L}(\boldsymbol{\theta}, x, y) =& \mathcal{L}_{sim}(O_{\boldsymbol{\theta}}(x, y), y) + \\
    &\quad\mathcal{L}_{const}(F_{\boldsymbol{\theta}}(x, y) - \alpha x, 0, \beta).
    \end{aligned}
    \label{eq:final-ob}
\end{equation}
Note that this objective is differentiable almost everywhere and can be optimized efficiently using stochastic gradient descent. 
Our method is summarized in Algorithm~\ref{alg:const-gen}, with
hyper-parameters and implementation details provided in the supplementary material.

{\renewcommand{\tabcolsep}{3pt}
\begin{figure*}
   \centering
\begin{tabular}{c@{}c|c@{}c@{}c|c}
Input & Ground Truth & From Scratch & Finetuned & Heuristic & Ours  \\
% \hline\hline

\includegraphics[width=0.16\textwidth]{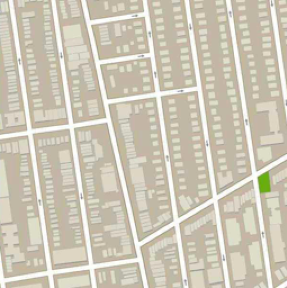}&
\includegraphics[width=0.16\textwidth]{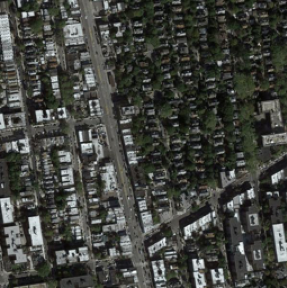}&
\includegraphics[width=0.16\textwidth]{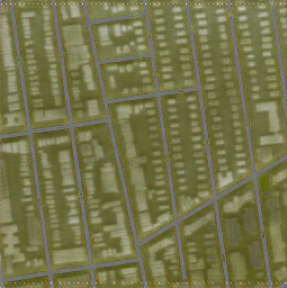}&
\includegraphics[width=0.16\textwidth]{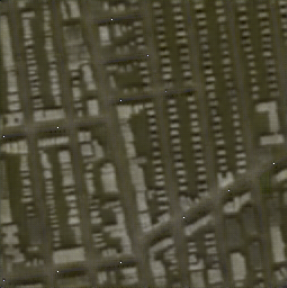}&
\includegraphics[width=0.16\textwidth]{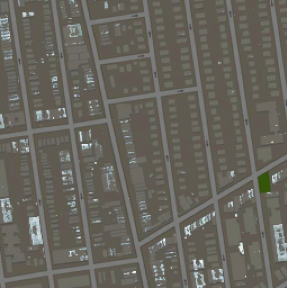}&
\includegraphics[width=0.16\textwidth]{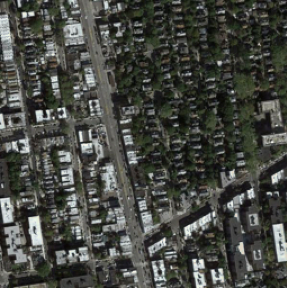}\\

\includegraphics[width=0.16\textwidth]{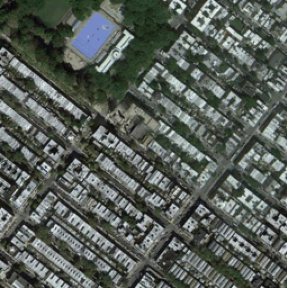}&
\includegraphics[width=0.16\textwidth]{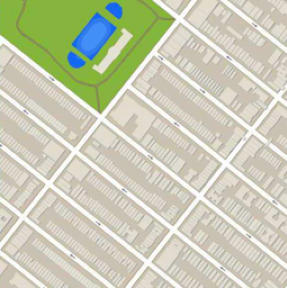}&
\includegraphics[width=0.16\textwidth]{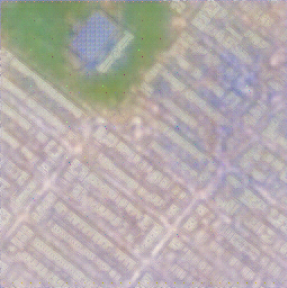}&
\includegraphics[width=0.16\textwidth]{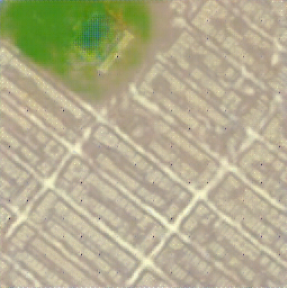}&
\includegraphics[width=0.16\textwidth]{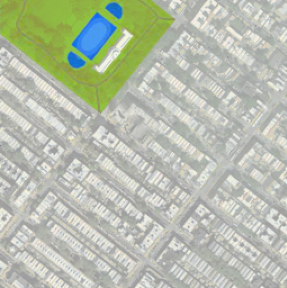}&
\includegraphics[width=0.16\textwidth]{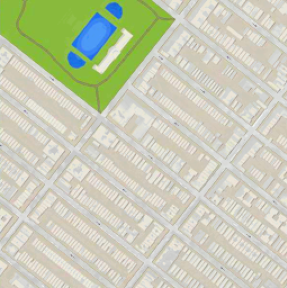}\\

\includegraphics[width=0.16\textwidth]{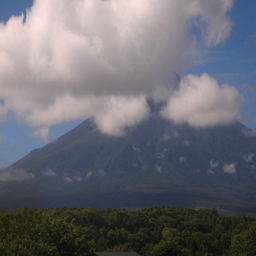}&
\includegraphics[width=0.16\textwidth]{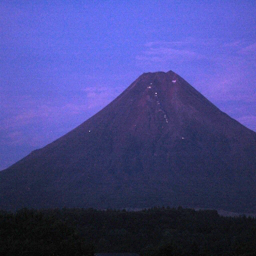}&
\includegraphics[width=0.16\textwidth]{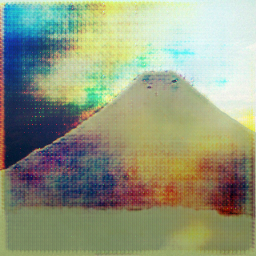}&
\includegraphics[width=0.16\textwidth]{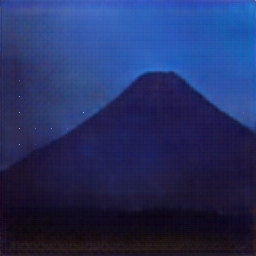}&
\includegraphics[width=0.16\textwidth]{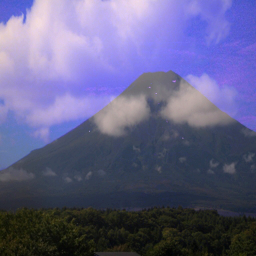}&
\includegraphics[width=0.16\textwidth]{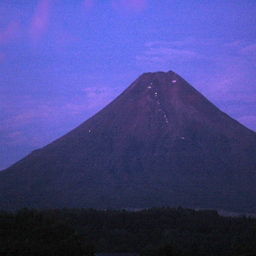}\\
\end{tabular}

\caption{Visual comparison to the baseline methods. Rows from top to bottom: Map$\rightarrow$Satellite, Satellite$\rightarrow$Map, Day$\rightarrow$Night}
\label{fig:qual} % I can do without the label too
\end{figure*}
}

\section{Experiments}\label{sec:experiments}

In this section, we begin by empirically evaluating 
%the efficacy of
our proposed method in comparison to baselines in a variety of tasks in Section~\ref{sec:expr-comparison} and Section~\ref{sec:expr-applications}.
We also conduct an ablation study on major design choices for our model (Section~\ref{sec:expr-ablation}), and provide an analysis of several interesting properties of our method (Section~\ref{sec:expr-properties}).
%Implementation details and hyper-parameters will be presented in the supplementary material.
We provide our code at \url{https://github.com/katieluo88/StayPositive}.

\paragraph{Datasets.} 
Most of our experiments are conducted on datasets from the image translation and image editing literature, which include semantic label$\leftrightarrow$photo \cite{isola2017image,zhu2017unpaired}, image$\rightarrow$style \cite{huang2017adain,zhu2017unpaired}, dog$\rightarrow$cat~\cite{Huang_2018_ECCV}, and CelebA-HQ dataset \cite{shen2020interpreting,sheng2018avatar,progressivegan}.
For unaligned datasets, we employ the provided pretrained models to generate a set of targets $y$.
If the dataset is paired, we will then use the ground truth images as the proposed image.

\paragraph{Metrics.} Following prior works \cite{kim2019ugatit,Liu2019FewShotUI,Choi2018StarGANUG,isola2017image}, we will use Frechet Inception Distance (FID)~\cite{heusel2017gans}, Kernel Inception Distance (KID)~\cite{alami2018unsupervised}, and Peak signal-to-noise ratio (PSNR).
KID computes the squared Maximum Mean Discrepancy of features extracted from Inception Network~\cite{szegedy2016rethinking} while FID computes the Frechet distance from those features.
At the same time, we also report the PSNR to quantify the similarity between the proposed images and the final output image, after normalization.
All reported numbers are computed by averaging across multiple $\alpha$ values.

\begin{figure}[ht]
\centering
\includegraphics[width=\linewidth,trim={0.45cm 0.5cm 0.8cm 0.5cm},clip]{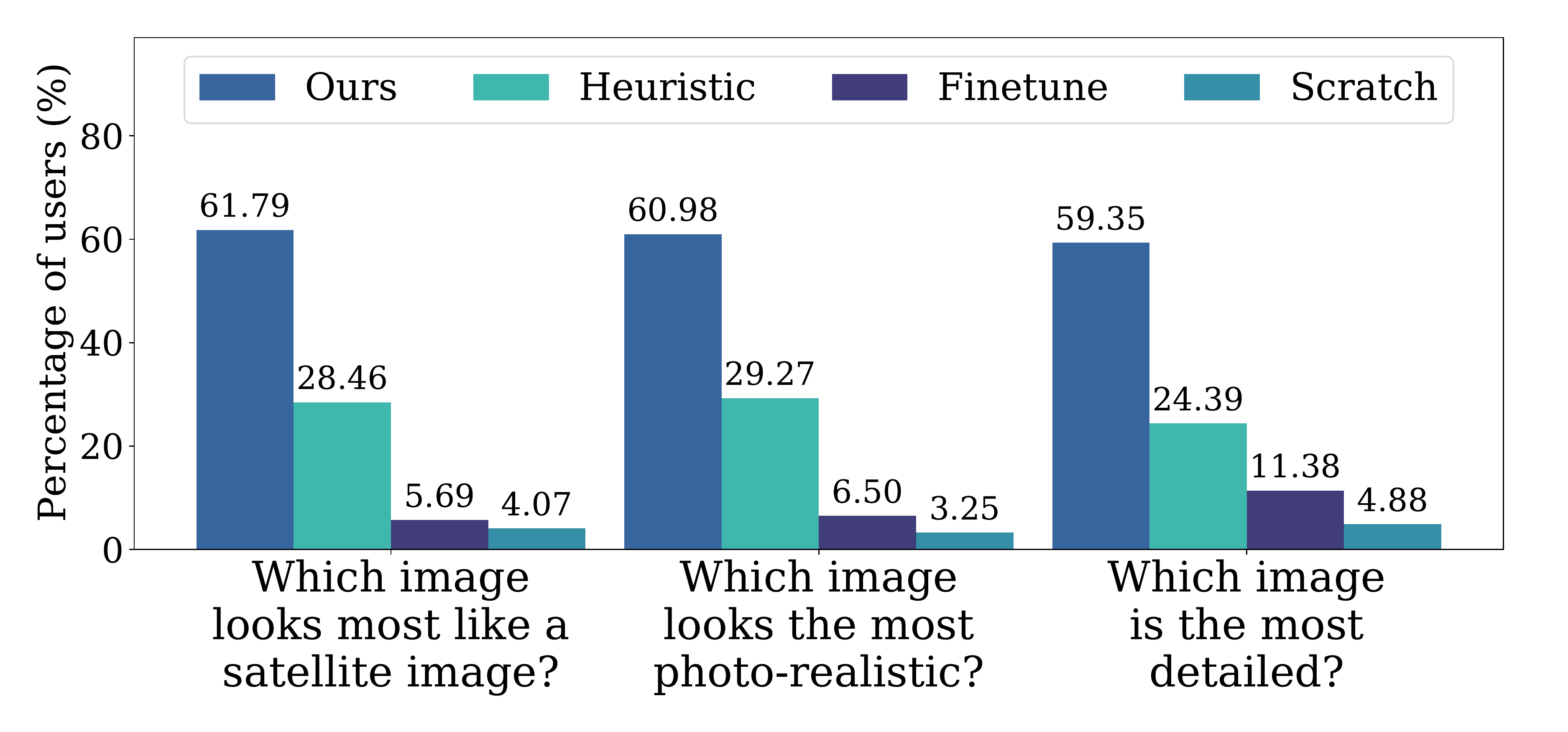}
\caption{Results from user study. We use AMT to recruit users to answer three questions asking for their preference on ``satellite image", ``photo-realistic", and ``detail". Most users prefer the output of our model over other baselines by a large margin.}
\label{fig:user-study-results}
\end{figure}
\subsection{Comparisons}\label{sec:expr-comparison}

%In this section, we want to compare our model with simple baselines built with state-of-the-arts models to show the efficacy of our method.
We consider the following three baselines.
The first one clips the difference required to reached the proposed image before adding to the input images to ensure physical feasibility (Heuristic).
The second one trains a Pix2pix~\cite{isola2017image} model from scratch to predict non-negative residuals used to predict the output image (From Scratch).
The third baseline uses our proposed loss to fine-tune a Pix2Pix generator (Finetune).
We train these three baselines and our methods on three tasks, including generating synthetic images (satellite$\rightarrow$map), generating photorealistic images (map$\rightarrow$satellite), and generating images with less light (day$\rightarrow$night).
The results are shown in Table~\ref{tab:baseline} and Figure~\ref{fig:qual}.
Our models out-perform baselines in FID in all tasks, and all but one task in KID and PSNR.
Even though our model does not lead in these metrics in the task of Satellite$\rightarrow$map, we can see from Figure~\ref{fig:qual} that our methods produce images with fewer ghosting artifacts for all three tasks.
In addition to the above mentioned metrics, we also conducted a user studies using Amazon Mechanical Turk. 
The results (Figure~\ref{fig:user-study-results}) suggest that most users chose images by our method over those from the baselines by a large margin.

{
\renewcommand{\tabcolsep}{2pt}
\begin{figure*}
\centering
\begin{tabular}{cccccc}
Day $\to$ Night~\cite{Anokhin_2020_CVPR} & 
Photo $\to$ Style~\cite{huang2017adain} & 
Style $\to$ Photo~\cite{huang2017adain} & 
Face $\to$ Smile~\cite{shen2020interpreting}&
% Dog $\to$ Cat~\cite{Huang_2018_ECCV} &
% Dog $\to$ Cat~\cite{Huang_2018_ECCV}
\multicolumn{2}{c}{Dog $\to$ Cats~\cite{Huang_2018_ECCV}}
\\
\includegraphics[width=0.16\textwidth]{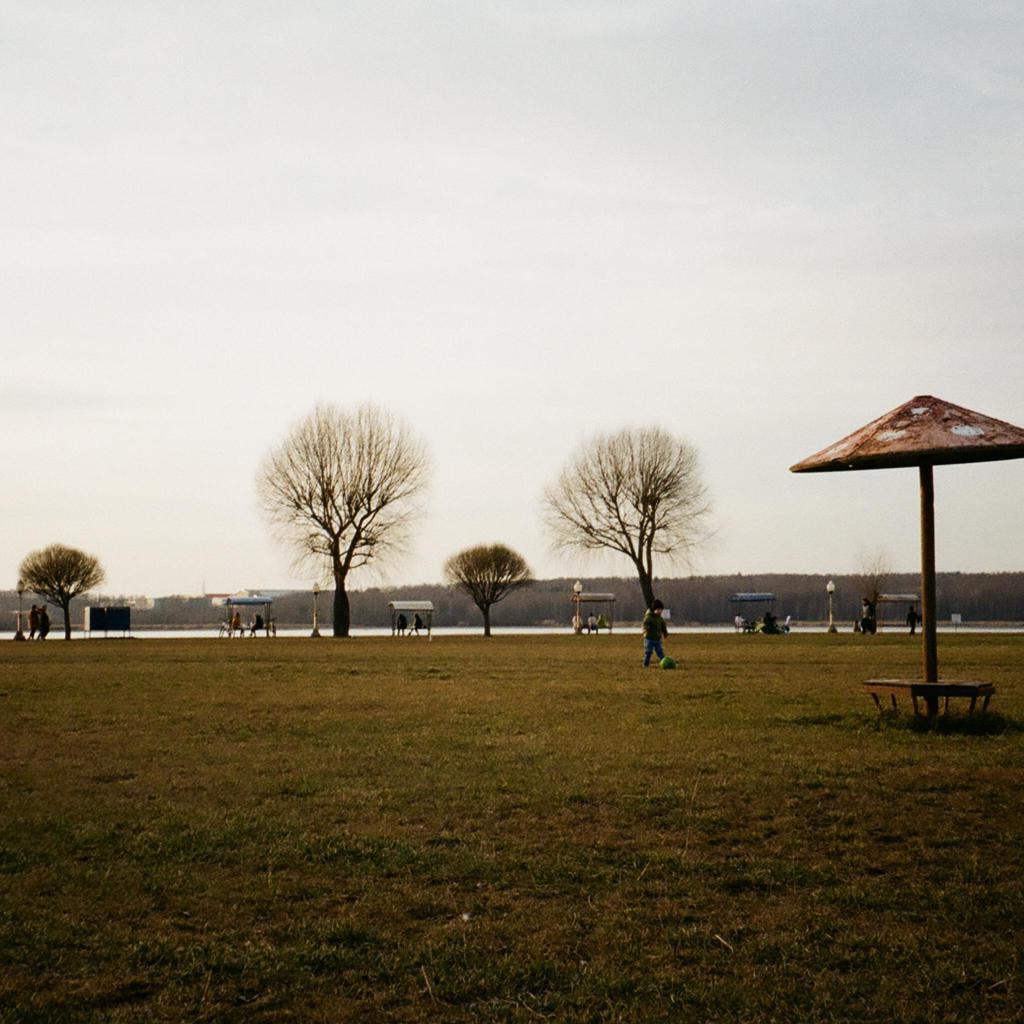}\llap{\includegraphics[height=1cm]{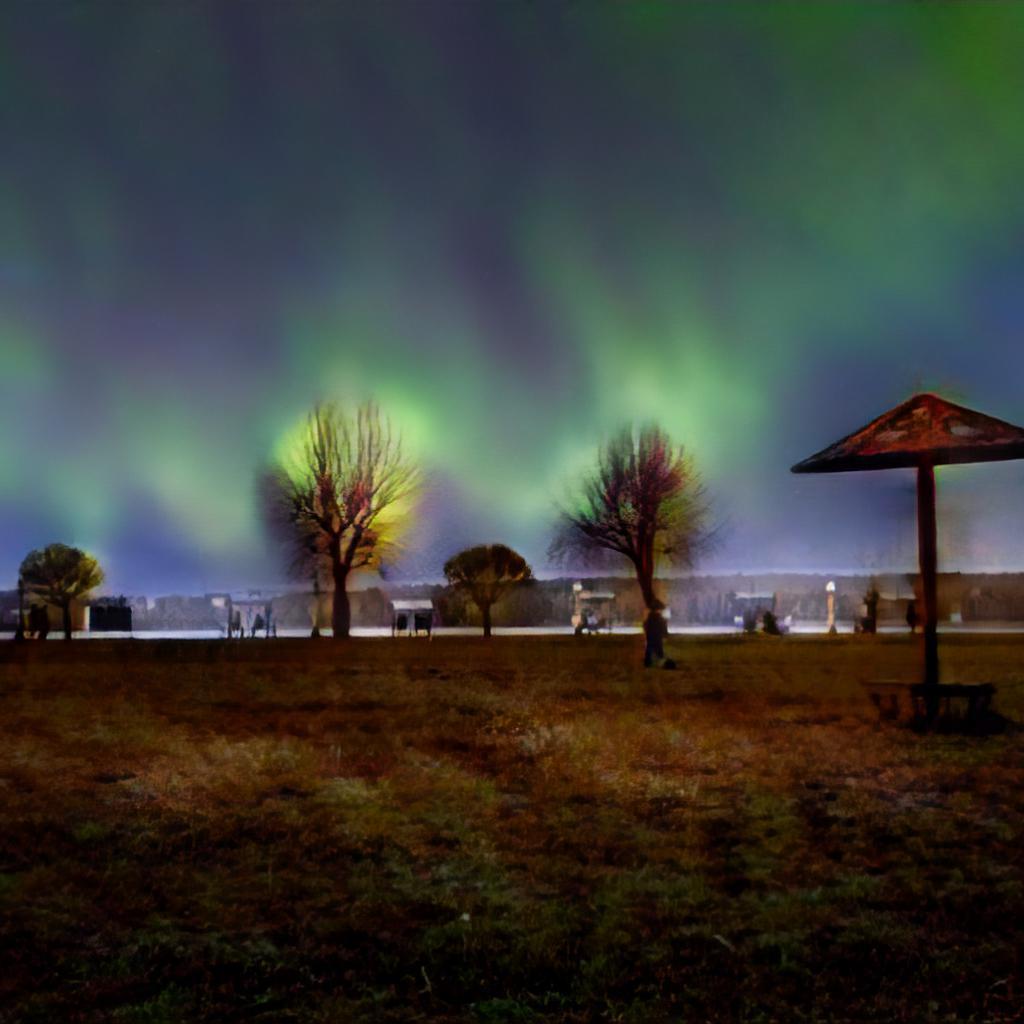}}&
\includegraphics[width=0.16\textwidth]{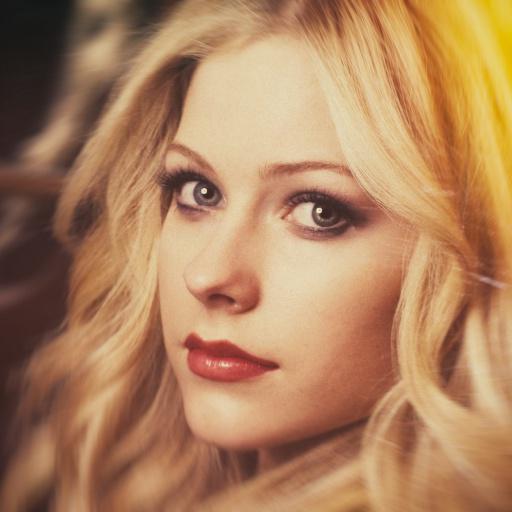}\llap{\includegraphics[height=1cm]{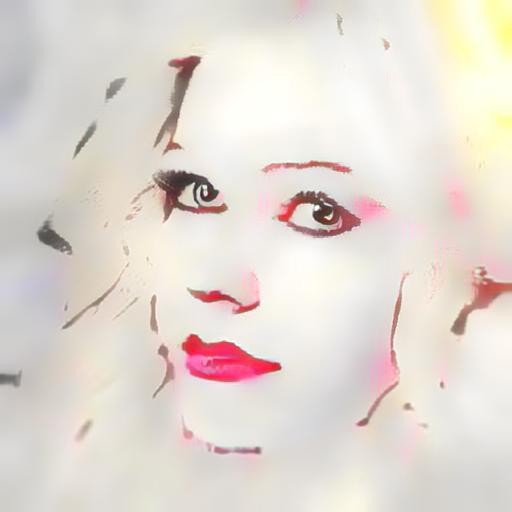}}&
\includegraphics[width=0.16\textwidth]{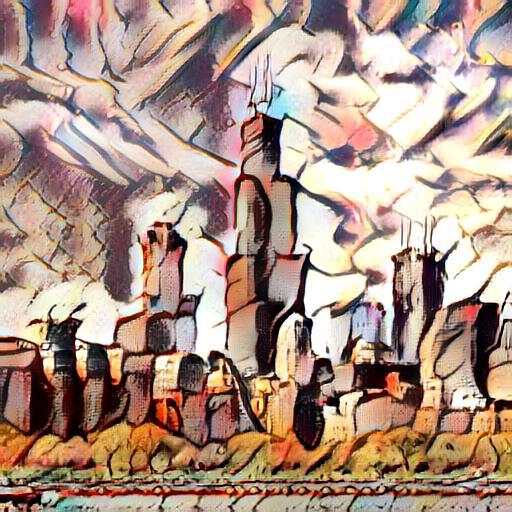}\llap{\includegraphics[height=1cm]{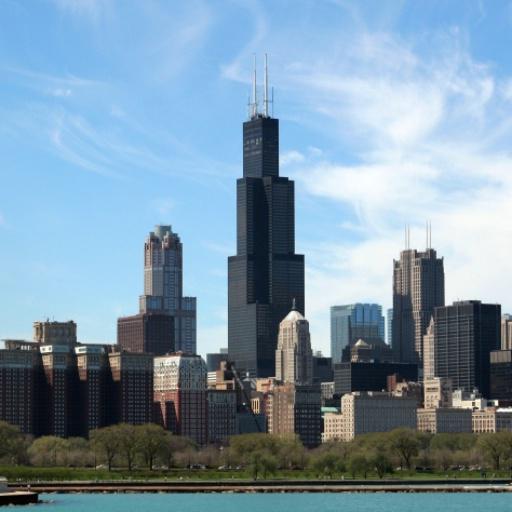}}&
\includegraphics[width=0.16\textwidth]{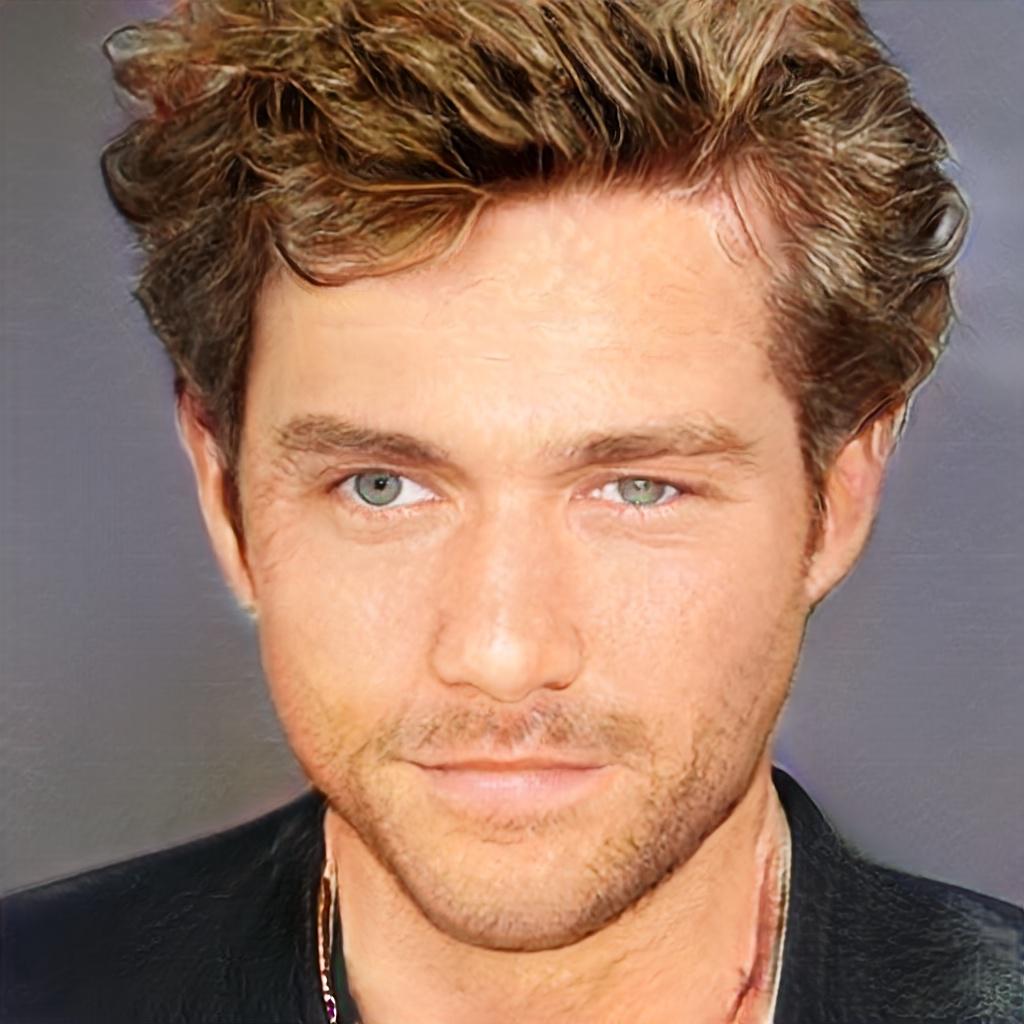}\llap{\includegraphics[height=1cm]{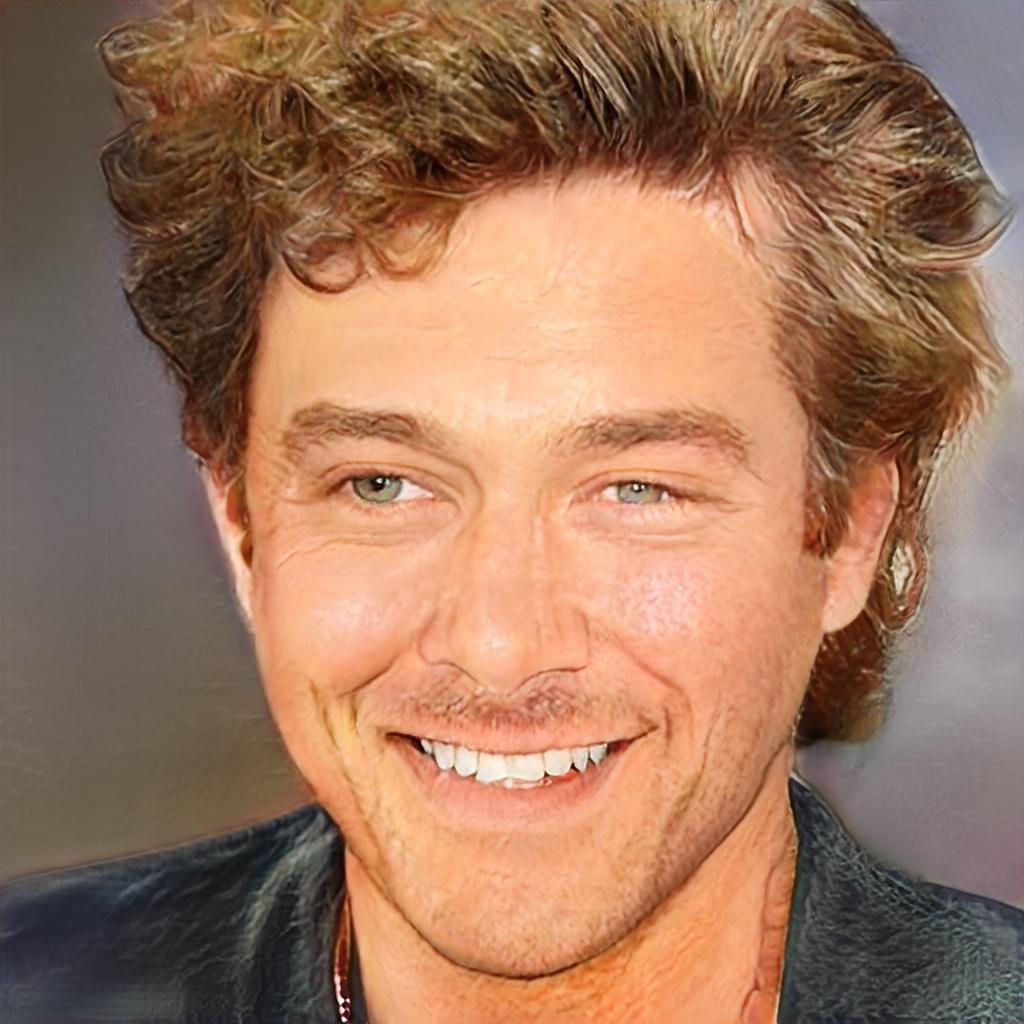}}&
\includegraphics[width=0.16\textwidth]{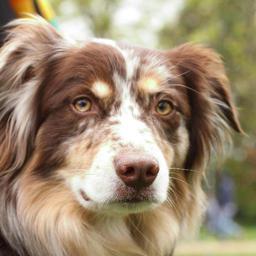}\llap{\includegraphics[height=1cm]{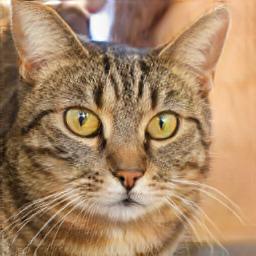}}&
\includegraphics[width=0.16\textwidth]{figures/tasks_comparison/dog2cat/dog2cat-input.jpg}\llap{\includegraphics[height=1cm]{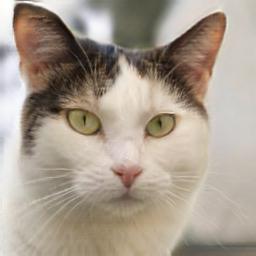}}
\\

\includegraphics[width=0.16\textwidth]{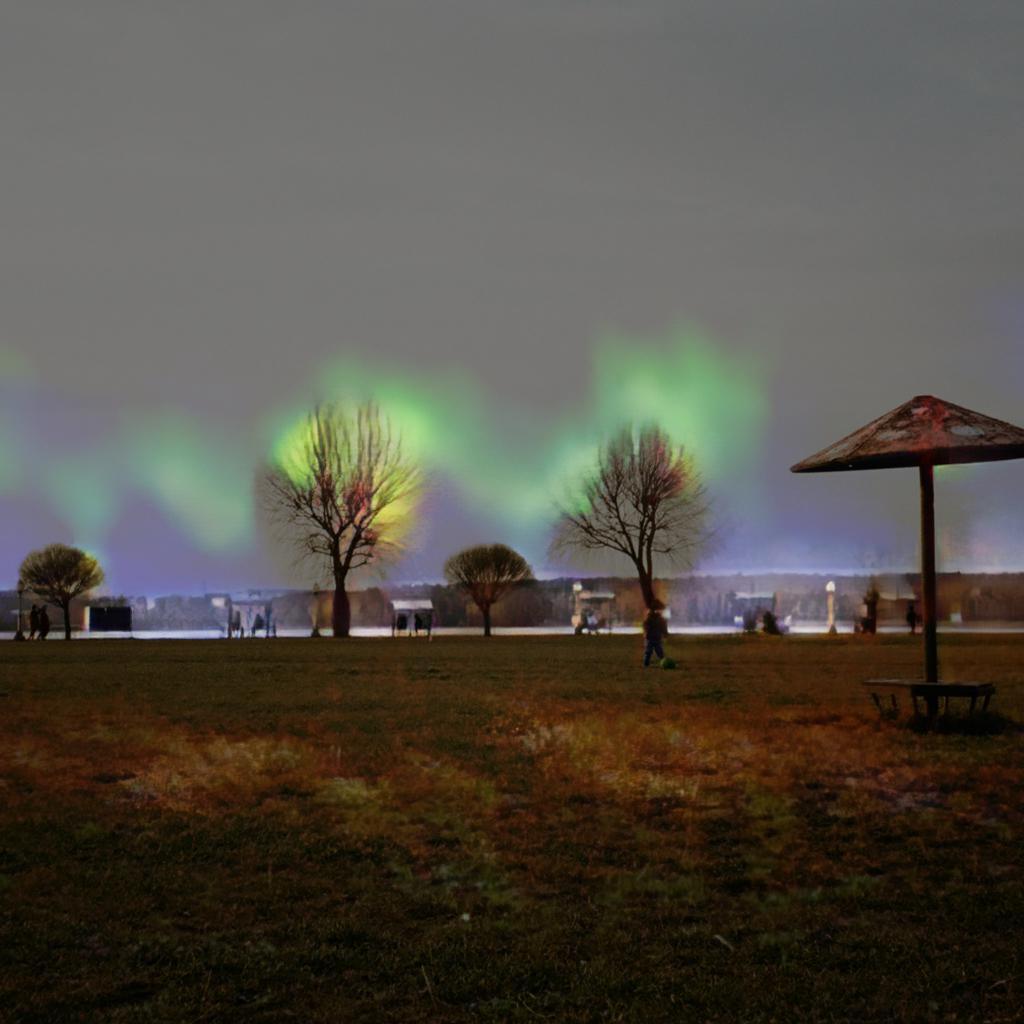}&
\includegraphics[width=0.16\textwidth]{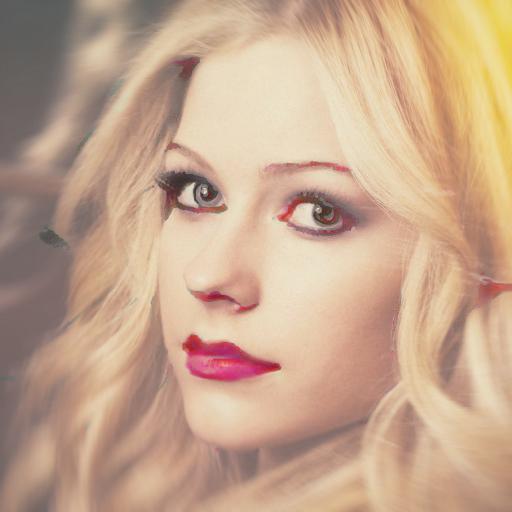}&
\includegraphics[width=0.16\textwidth]{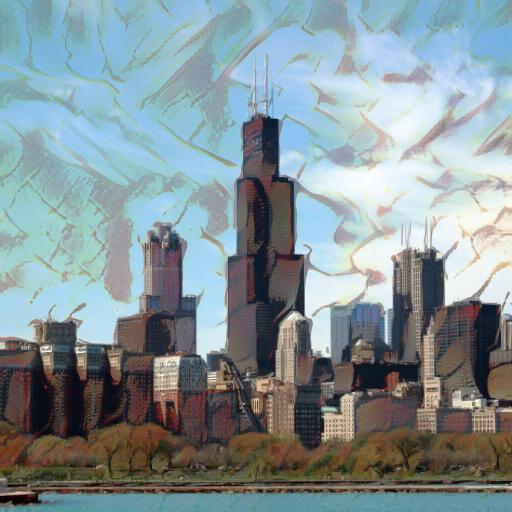}&
\includegraphics[width=0.16\textwidth]{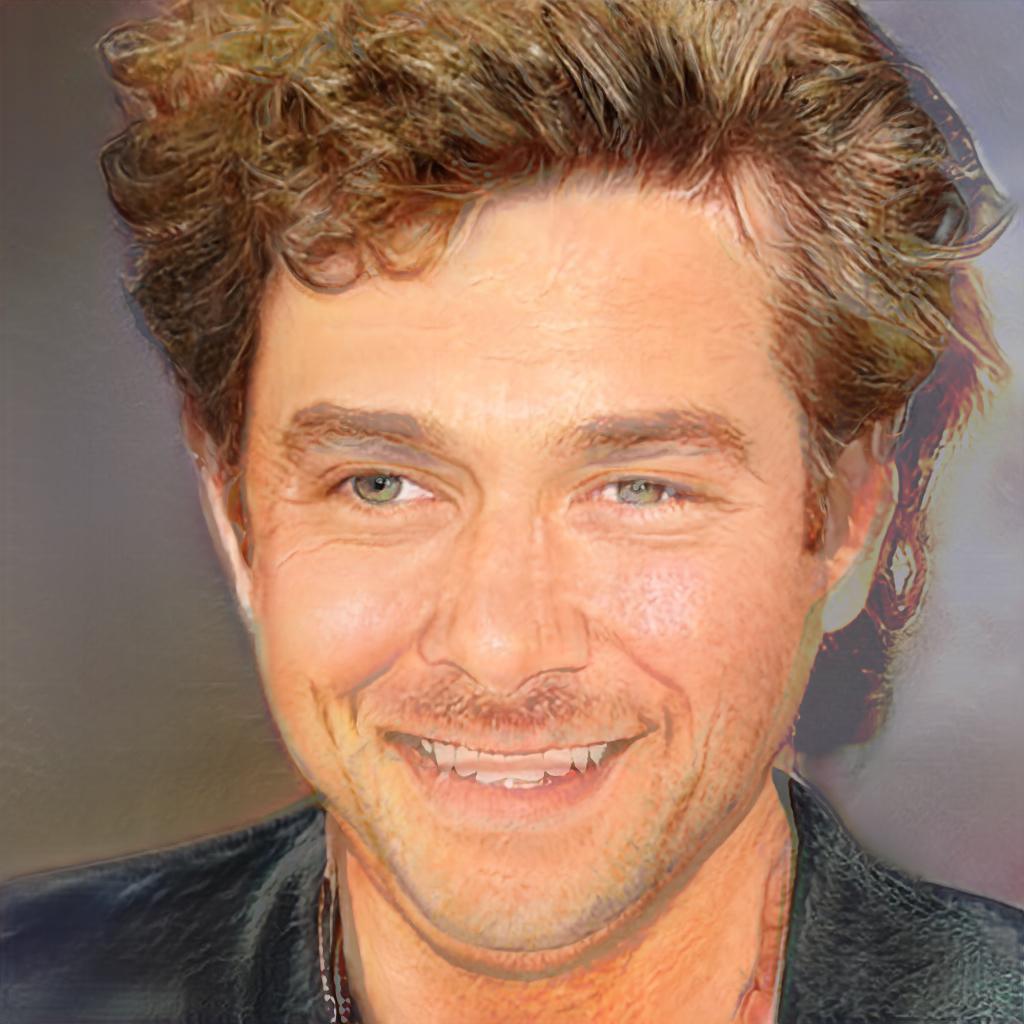} &
\includegraphics[width=0.16\textwidth]{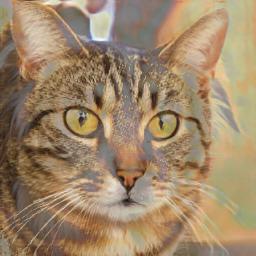}&
\includegraphics[width=0.16\textwidth]{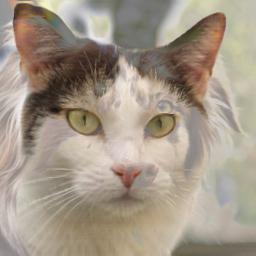}
\\

\includegraphics[width=0.16\textwidth]{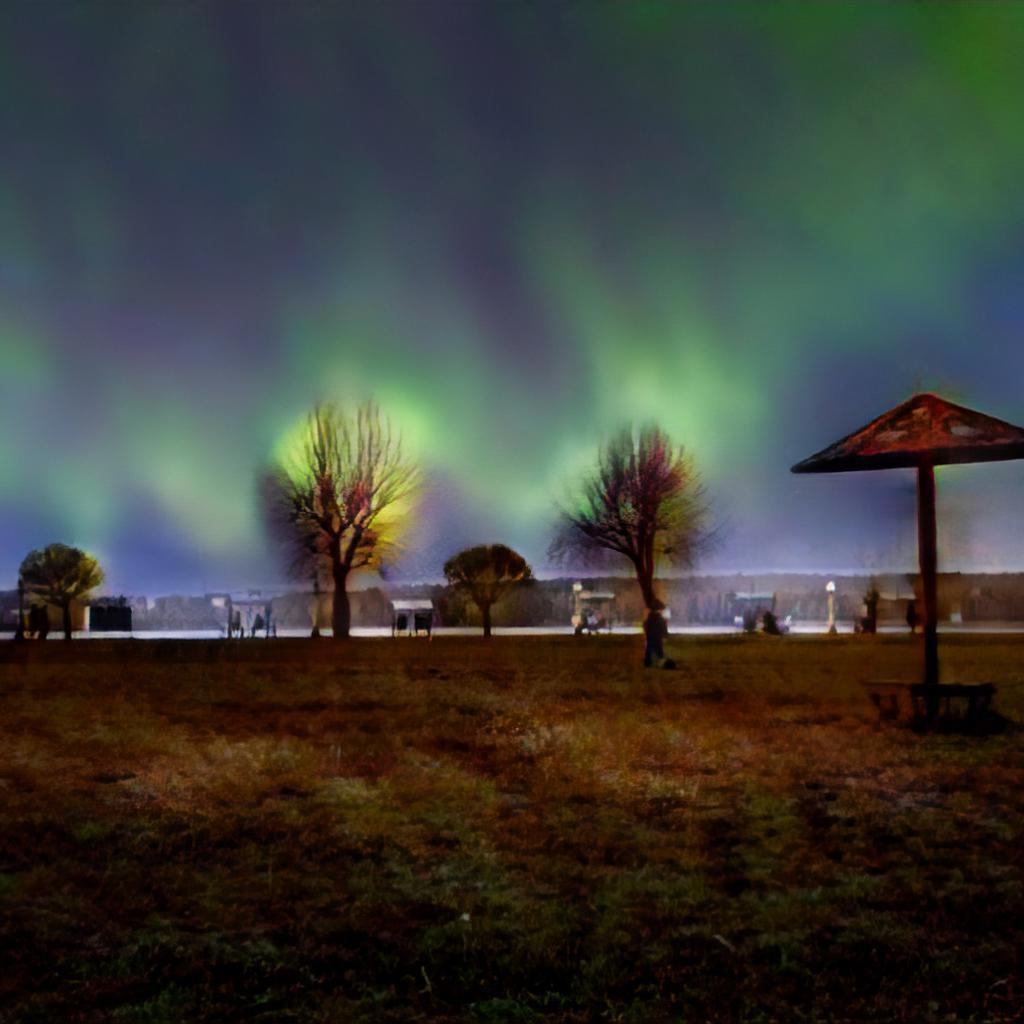}&
\includegraphics[width=0.16\textwidth]{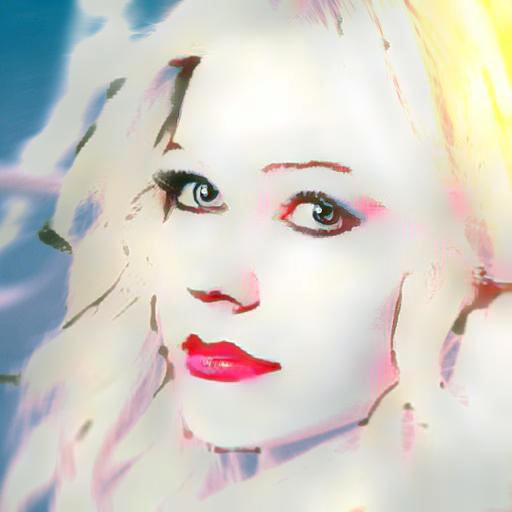}&
\includegraphics[width=0.16\textwidth]{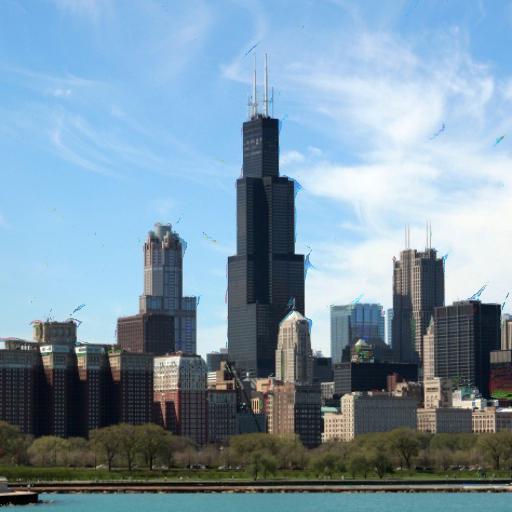}&
\includegraphics[width=0.16\textwidth]{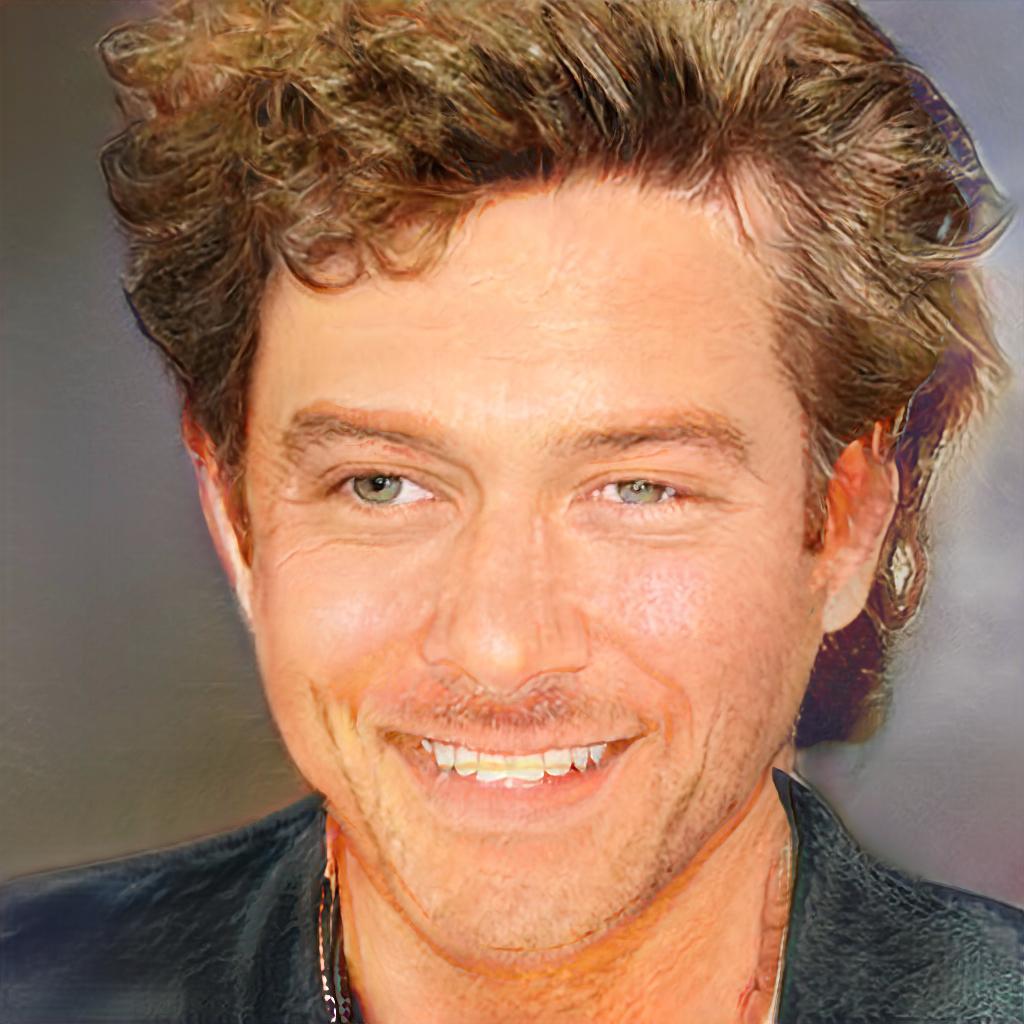}&
\includegraphics[width=0.16\textwidth]{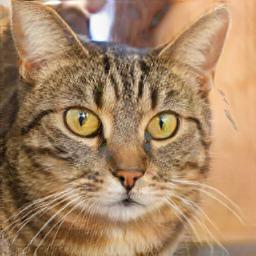}&
\includegraphics[width=0.16\textwidth]{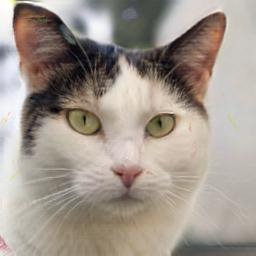}
\\

\end{tabular}
\caption{We compare our results across a variety of domains. From top to bottom: input (with proposal overlaid), heuristic baseline, our method. Observe that with our framework, we are able to obtain multimodal outputs from the generative model.}
\label{fig:resid} % I can do without the label too
\end{figure*}
}

\begin{table}[t]
\begin{center}
\begin{tabular}{@{}lcc@{}}
\toprule
Datasets                       & Baseline & Ours \\
\midrule
Zebra $\rightarrow$ Horse    & 12.63    & \textbf{11.53}      \\
Winter $\rightarrow$ Summer  & 1.16     & \textbf{1.03}       \\
Photo $\rightarrow$ Monet    & 1.59     & \textbf{1.38}       \\
Photo $\rightarrow$ Von Gogh & 3.27     & \textbf{3.26}       \\
Photo $\rightarrow$ Cezanne  & 4.70     & \textbf{4.51}       \\
Photo $\rightarrow$ Uyeoko   & 10.33    & \textbf{9.71}       \\ \bottomrule
\end{tabular}
\end{center}
\caption{Kernel Inception Distance$\times 100$ metric on \cite{zhu2017unpaired}. Our model outperforms the heuristic baseline on a wide range of tasks.}
\label{tab:tasks_baseline}
\end{table}

% \input{figures/tasks_compare}

% \paragraph{User study}
% While a variety of metrics have confirmed the efficacy of our proposed method, these metrics were designed to only approximate what a real user's preference is.
% In order to get a more accurate assessment of whether users would prefer the output of our method to the baselines, we conducted a user study using Amazon Machanical Turks on the Map$\rightarrow$Satellite dataset.
% For each selected input image, we will presented the four output images generated by four methods in random order in a row, and asked the workers four questions:
% \begin{enumerate}
%     \item Which image looks most like a satellite image?
%     \item Which image looks the most photo-realistic?
%     \item Which image is the most detailed?
%     \item Which image looks most like a satellite map?
% \end{enumerate}
% As we can see from the data, the user prefer our model to any other baseline by a large margin.

\subsection{Applications} \label{sec:expr-applications}
As mentioned in Section~\ref{sec:two-step}, our method can be combined with different image translation methods.
In this section, we quantitatively compare our method with the heuristic baseline on a set of unpaired image translation datasets listed in Table~\ref{tab:tasks_baseline}, adapted from CycleGAN~\cite{zhu2017unpaired}.
The table suggests that our model achieves comparable or better KID than the heuristic baseline in all tasks.
We also apply our method in a variety of more difficult datasets such as high resolution day$\rightarrow$night~\cite{Anokhin_2020_CVPR}, style transfer~\cite{huang2017adain}, sketch to photo~\cite{huang2017adain}, dog$\rightarrow$cat~\cite{Huang_2018_ECCV}, and face attribute editing~\cite{shen2020interpreting,shen2020interfacegan}.
The qualitative examples of aforementioned tasks are presented in Figure~\ref{fig:resid}.
While the images produced by the heuristic based method tend to exhibit ghosting artifacts arising from enforcing non-negativity, 
our method is able to avoid such shortcomings and produce high quality images that can resemble the target.

\subsection{Analysis}\label{sec:expr-analysis}

In this section, we will first conduct 
%a detailed 
an
ablation study to examine our design decisions.
Then we will present an analysis of how our methods perform according to different $\alpha$ values.
Finally, we will use qualitative examples to illustrate how our proposed approach takes advantage of human perceptual quirks.

\begin{table}[t]
\begin{center}
\begin{tabular}{@{}lccc@{}}
\toprule
Method    & FID($\downarrow$)    & KID($\downarrow$)   & N-PSNR($\uparrow$) \\
\midrule
w/o normalization  & 146.95 & 13.23 & 45.86  \\
w/o $\mathcal{L}_{const}$ & 139.95 & 13.74 & 69.09  \\
w/o $\mathcal{L}_{sim}$    & 145.21 & 14.57 & 65.71 \\ 
Per-pixel prediction      & 132.82 & 13.90 & \textbf{78.17}  \\ 
\midrule
Ours         & \textbf{125.75} & \textbf{12.76} & 68.90 \\ 
\bottomrule
\end{tabular}    
\end{center}
\caption{Ablation study on the Map$\rightarrow$Satellite task. We explore in this order: normalization methods, loss function importance, and our method's transformation groups.}
\label{tab:ablations}
\end{table}
\paragraph{Ablations.}\label{sec:expr-ablation}
We evaluate three design decisions: normalizing the dynamic range before computing the perceptual similarity $\mathcal{L}_{sim}$ (Section~\ref{sec:reconloss}), minimizing both perceptual similarity $\mathcal{L}_{sim}$ and constraint violation $\mathcal{L}_{const}$  (Section~\ref{sec:resloss}), and predicting global modification instead of per-pixel modification (Section~\ref{sec:offset}).
We run ablations of our method on the task map$\rightarrow$satellite and reported results in Table~\ref{tab:ablations}.
It's interesting to note that predicting per-pixel modification achieves better PSNR score, which is expected due to the large amount of parameters, but produces worse FID and KID, which suggests that the method creates images that are semantically far away from the target domain.
On the other hand, our methods achieve the best FID and KID among all variants.

\paragraph{Effect of varying $\alpha$.}

% \begin{figure}[t]
%     \centering
%     \includegraphics[width=\linewidth,trim={0.73cm 0.6cm 0.5cm 0.5},clip]{figures/alpha_plot.png}
%     \caption{KID and FID metrics on our method vs. heuristic baseline. Our methods outperforms baselines across almost all $\alpha$.}
%     \label{fig:alpha-plot}
% \end{figure}
\begin{figure}[t]
    \centering
    \includegraphics[width=\linewidth,trim={0 0.5cm 0 0.5},clip]{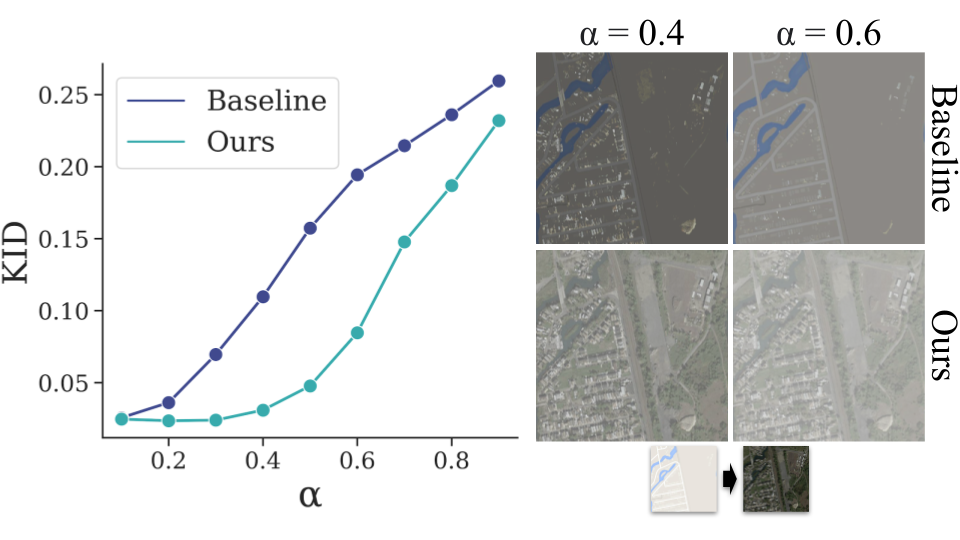}
    \caption{Results on a range of $\alpha$ values for Map$\rightarrow$Satellite. We plot KID metrics on our method vs. heuristic baseline. Our methods outperforms baselines across almost all $\alpha$. 
    Visually, our method also produces less artifacts comparing to the baseline.
    }
    \label{fig:alpha-plot}
\end{figure}

% \begin{figure}[t]
%     \centering
%     \includegraphics[width=\linewidth,trim={0 1cm 0 2cm},clip]{figures/alpha_compare.png}
%     \caption{Visual result on a range of $\alpha$ values for Map$\rightarrow$Satellite. Our method does better at higher $\alpha$ than the heuristic baseline.}
%     \label{fig:alpha-compare}
% \end{figure}

While previous experiments show that our methods outperforms the baselines when averaging across different values of $\alpha$, it's thus far unclear how our method compares to the baselines at different $\alpha$ values.
In Figure~\ref{fig:alpha-plot}, we present the change of KID across 10 different $\alpha$ values between our methods and the heuristic baseline.
When $\alpha$ is small, both the baseline and our method perform well, as expected, since the task is very similar to the one without non-negativity constraint.
On the other hand, when $\alpha$ is large, both our method and the baseline tend to perform badly (though our method still out-performs the baseline), which is expected since the task is almost impossible.
It's interesting to note that our method is able to maintain good performance until after $\alpha > 0.5$, while the baseline's performance degrades drastically.
This suggests that our method can produce images in harmony with more ambient light.
Visually, the ghosting artifacts from the input image does not show up in our method until higher $\alpha$ values.
% A similar trend is evident in Figure~\ref{fig:alpha-compare} \sjb{which figure is referenced here?}, as the ghosting artifact from the input images does not show up in our method until $\alpha$ reaches $0.8$.

\subsection{Leveraging quirks of human perception}\label{sec:expr-properties}

%Is our model able to leverage human visual quirks such as lightness constancy or insensitivity to high texture details as it's designed to?
In this section we present qualitative examples to demonstrate how our model benefits from quirks of human visual perception involving.

% {
% \renewcommand{\tabcolsep}{1pt}
% \begin{figure}
% \begin{center}
% \includegraphics[width=\linewidth]{figures/white_balance_tricks/wb.pdf}
% \end{center}
% \caption{\todo{Claim 11: our model with normalization can play the white balancing trick to produce dark pixel (go back to the intro)}}
% \label{fig:whilte-balance-example}
% \end{figure}
% }

{
\renewcommand{\tabcolsep}{1pt}
\begin{figure}
\begin{center}
\begin{tabular}{ccc}
Input & 
Output & 
Target \\
\includegraphics[width=0.33\linewidth]{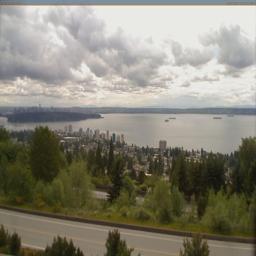}&
\includegraphics[width=0.33\linewidth]{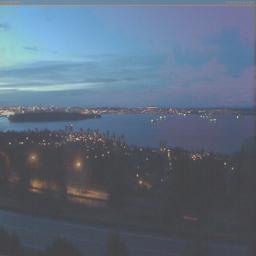}&
\includegraphics[width=0.33\linewidth]{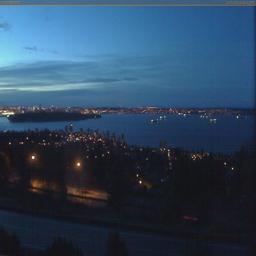}
\\
\includegraphics[width=0.33\linewidth]{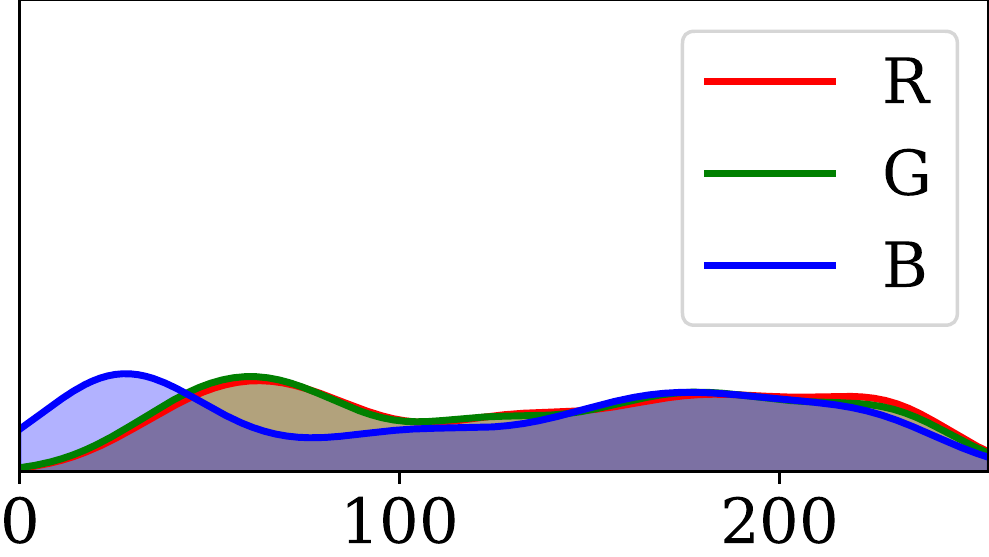}&
\includegraphics[width=0.33\linewidth]{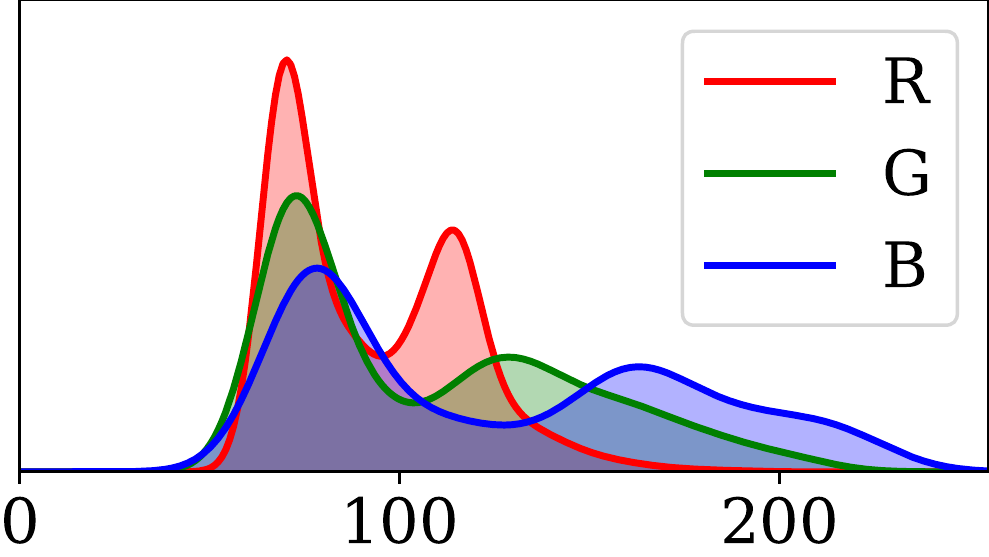}&
\includegraphics[width=0.33\linewidth]{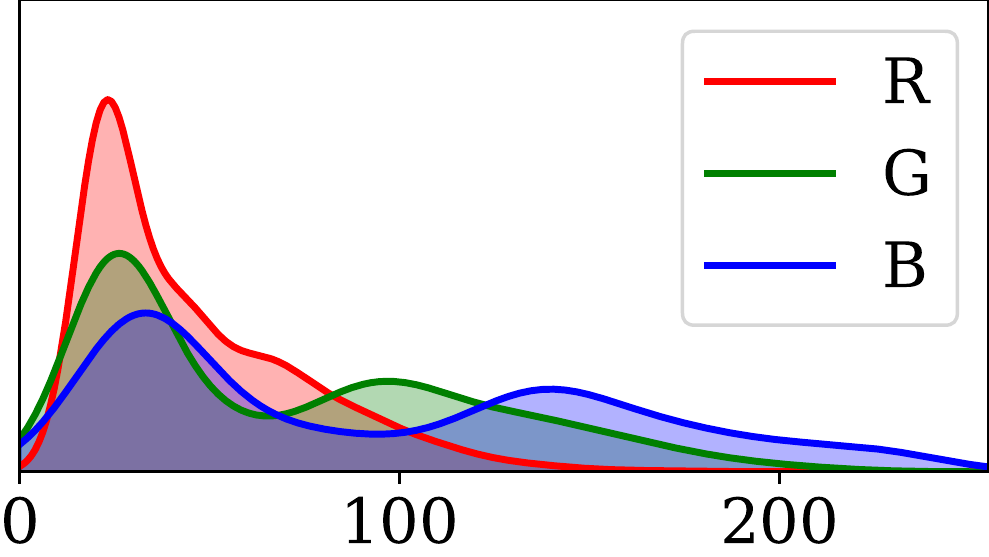}
\end{tabular}
\end{center}
\caption{
Demonstration that our model can leverage the \textit{lightness constancy} produce the appearance of dark pixels through the addition of light.
}
\label{fig:white-balance-example}
\end{figure}
}
\paragraph{Lightness constancy.}
In Figure~\ref{fig:white-balance-example}, we show how our model turns a daytime image into a nighttime image, along with the associated color histogram of the input and target images. 
Note that such a task is infeasible, at face value, since the nighttime image contains more pixels with a lower amount of blue or red light compared to the input image, as shown in the histogram.
Our model is nonetheless capable of producing an image that looks similar to the target image without subtracting any light.
The model achieves this result by padding the amount of light around the points in need of darkening, which shifts the whole dynamic range higher and brighter (i.e., the histogram of the output image has few pixels with value less than $50$).
Though the ``black" point in the middle of the image has RGB values greater than $50$, such a change is generally hard to detect without direct, side-by-side comparison to the target image.

{
\renewcommand{\tabcolsep}{1pt}
\begin{figure}
\begin{center}
\begin{tabular}{ccc}
Input & 
Output & 
Target \\
\includegraphics[width=0.33\linewidth]{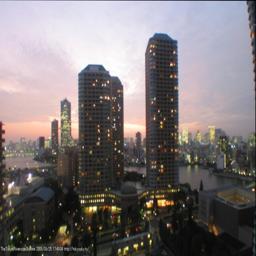}&
\includegraphics[width=0.33\linewidth]{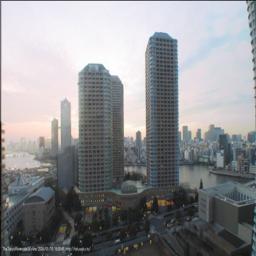}&
\includegraphics[width=0.33\linewidth]{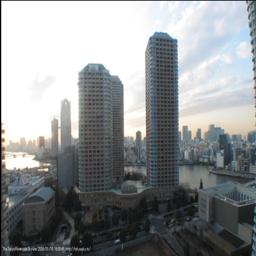}\\
\includegraphics[width=0.33\linewidth]{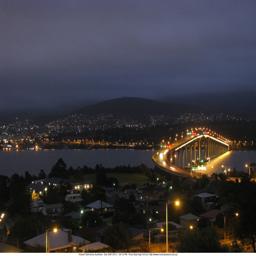}&
\includegraphics[width=0.33\linewidth]{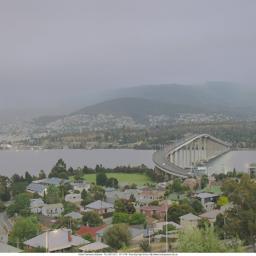}&
\includegraphics[width=0.33\linewidth]{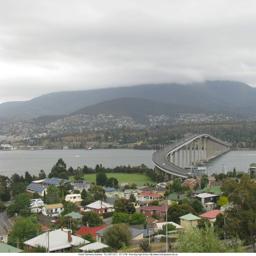}\\
\end{tabular}
\end{center}
\caption{
% \todo{
Our model can hide mistakes and non-salient image details. At first glance, the output image is a normal day photo. The building lights that are left on are details ignored by our method.}
% }
\label{fig:hide_in_details}
\end{figure}
}
\paragraph{Insensitivity to errors in textured regions.}
Figure~\ref{fig:hide_in_details} shows examples of our method turning nighttime images to daytime images of city scenes.
The challenge in such tasks lies in ``turning off'' streetlamps or office window lights when subtraction is not allowed.
Interestingly, our model tends not to worry too much about turning off all the lights since the associated discrepancy is very small and usually does not attract users' attention.
While all of our output images appear natural at first glance, upon closer examination, these simulated daytime images retain evidence of turned-on lights.
However, our visual system is capable of handling these minor errors.

\section{Related Work}\label{sec:related-works}

Recent deep generative models such as flow-based model~\cite{flow,Kingma2018GlowGF,Dinh2017DensityEU}, auto-regressive models~\cite{oord2016pixel,Salimans2017PixelCNNIT,Parmar2018ImageT}, VAEs~\cite{vae,Kingma2017ImprovedVI,Maale2019BIVAAV,Vahdat2020NVAEAD}, and GANs~\cite{gan,dcgan,wgan,iwgan,karras2019style,karras2019analyzing,biggan,progressivegan} have contributed to improvements in unconditional image generation.
The literature on conditional image generation explores related tasks including  
style transfer~\cite{gatys2016image,ulyanov2017improved,chen2016fast,huang2017adain,Dumoulin2017ALR,Anokhin_2020_CVPR,Bahng2020ExploringUF,Park2019ArbitraryST}, 
supervised image translation~\cite{isola2017image,Zhu2017TowardMI,Ghosh2018MultiagentDG,Lee2019HarmonizingML,Li2019DiverseIS,Lee2020MaskGANTD,Liu2019LearningTP,Park_2019_CVPR,Tang2020EdgeGG,Wang2018HighResolutionIS,Zheng2020ExampleGuidedIS},
unsupervised image translation~\cite{liu2017unsupervised,zhu2017unpaired,Huang_2018_ECCV,Benaim2017OneSidedUD,Choi2018StarGANUG,Liu2019FewShotUI,Saito2020COCOFUNITFU},
and semantic image editing~\cite{Shen2020InterpretingTL,bau2019gandissect}.
Please refer to \textit{Liu et al}~\cite{Liu2020GenerativeAN} for more details.
Despite this progress, the above mentioned works assume the ability to output any color at any pixel location, which is ill-suited to our problem setting.
In fact, our method can be combined with all of the above image translation models.
% Some prior works have studied
A related constrained image generation problem is studied by Heim~\cite{CONGAN}.
This work, however, focuses on semantic constraints, \eg, that the generated image should be closer to one attribute group than the other.
%Our present work sets up to tackle the image generation problem constrained on the physical rules created by the augmented reality setting.
Most closely related to our present work is the recent literature on projector compensation~\cite{Huang2020DeLTraDL,compenNetpp,projectorcomp,Boroomand2016SaliencyguidedPG,Asayama2018FabricatingDV,Harville2006PracticalMF,Siegl2015RealtimePL,Aliaga2012FastHA,Ashdown2006RobustCP}, where one updates the projected images to account for the potentially non-planar and non-uniformly colored projector surface.
% \sjb{I added ``non-uniformly colored,'' is there a better way to say this?}
Our work focuses instead on the OST setting, in which the image is formed via semi-transparent glass instead of at the projector surface.
In the sense that our our approach plays with the effective dynamic range and the contrast of the image, our work is related to \textit{tone mapping}, a process of mapping real world colors to a restricted color space for displaying image while preserving details.
Traditional tone mapping algorithms employ variants of histogram equalization~\cite{yoon2007image, choi2011contrast,duan2010tone, boschetti2010high, ploumis2016perception} to enhance the contrast of LDR images.
Many recent works~\cite{marnerides2018expandnet,deeptone,eilertsen2017hdr,liu2020single} leverage deep learning to recover the missing details in the over-exposed image regions by expanding the dynamic range of single LDR images.
In our paper, we found that pairing very simple contrast and dynamic range adjustment with SGD is sufficient to obtain our desired results.
% \todo{White balancing}

\section{Conclusion}

Through this work, we explored the problem of non-negative image generation and proposed an effective framework that achieves consistently better results than the baselines.
Possible extensions to this work include exploring additional properties of the human visual system and removing the overly simplistic modeling assumptions. 
Thus far, our studies have not included implementation using a real-world AR rig. We are pursuing the testing of our method in a physical system in an ongoing work.

\vspace{-10px}
\paragraph{Acknowledgements} This work was supported in part by grants from Magic Leap and Facebook AI, and a donation from NVIDIA.

\clearpage
{\small
\bibliographystyle{ieee_fullname}
\bibliography{main}
}

\end{document}